\documentclass{article}

\usepackage{microtype}
\usepackage{graphicx}
\usepackage{subfigure}
\usepackage{booktabs} 
\usepackage{setspace}
\usepackage{hyperref}

\usepackage[accepted]{icml2024}

\usepackage{amsmath}
\usepackage{amssymb}
\usepackage{mathtools}
\usepackage{amsthm}

\theoremstyle{plain}
\newtheorem{theorem}{Theorem}[section]

\newtheorem{lemma}[theorem]{Lemma}

\theoremstyle{definition}

\newtheorem{assumption}[theorem]{Assumption}
\theoremstyle{remark}

\usepackage[textsize=tiny]{todonotes}

\usepackage{mystyle}

\icmltitlerunning{Pessimism Meets Risk: Risk-Sensitive Offline Reinforcement Learning
}

\begin{document}

\twocolumn[
\icmltitle{Pessimism Meets Risk: Risk-Sensitive Offline Reinforcement Learning 
}



\icmlsetsymbol{equal}{*}

\begin{icmlauthorlist}
\icmlauthor{Dake Zhang}{aa}
\icmlauthor{Boxiang Lyu}{aa}
\icmlauthor{Shuang Qiu${}^\dag$}{bb}
\icmlauthor{Mladen Kolar}{cc}
\icmlauthor{Tong Zhang}{dd}
\end{icmlauthorlist}

\icmlaffiliation{aa}{University of Chicago, IL, USA}
\icmlaffiliation{bb}{Hong Kong University of Science and Technology, Hong Kong, China}
\icmlaffiliation{cc}{University of Southern California, CA, USA}
\icmlaffiliation{dd}{University of Illinois Urbana-Champaign, IL, USA  ${}^\dag$Corresponding Author}
\icmlcorrespondingauthor{Shuang Qiu}{masqiu@ust.hk}

\icmlkeywords{Machine Learning, ICML}

\vskip 0.3in
]



\printAffiliationsAndNotice{}  


\begin{abstract}
We study risk-sensitive reinforcement learning (RL), a crucial field due to its ability to enhance decision-making in scenarios where it is essential to manage uncertainty and minimize potential adverse outcomes. Particularly, our work focuses on applying the entropic risk measure to RL problems.  While existing literature primarily investigates the online setting, there remains a large gap in understanding how to efficiently derive a near-optimal policy based on this risk measure using only a pre-collected dataset. We center on the linear Markov Decision Process (MDP) setting, a well-regarded theoretical framework that has yet to be examined from a risk-sensitive standpoint. In response, we introduce two provably sample-efficient algorithms. We begin by presenting a risk-sensitive pessimistic value iteration algorithm, offering a tight analysis by leveraging the structure of the risk-sensitive performance measure. To further improve the obtained bounds, we propose another pessimistic algorithm that utilizes variance information and reference-advantage decomposition, effectively improving both the dependence on the space dimension $d$ and the risk-sensitivity factor. To the best of our knowledge, we obtain the first provably efficient risk-sensitive offline RL algorithms. 
\end{abstract}

\addtocontents{toc}{\protect\setcounter{tocdepth}{-1}}   

  \vspace{-0.1cm}
\section{Introduction}

Reinforcement learning (RL) with risk-sensitivity is becoming increasingly popular in a variety of real-world risk-sensitive problems, such as finance~\citep{follmer2002convex,hambly2021recent}, optimal control~\citep{nass2019entropic}, 
and neuroscience and psychology~\citep{chateauneuf1994risk,nagengast2010risk,braun2011risk,niv2012neural}.
Whereas researchers have been focusing on analyzing and understanding how a risk-sensitive near-optimal policy can be learned in the online setting \citep{fei2021exponential, fei2021risk, fei2020risk, fei2022cascaded,liang2022bridging,du2023provably, wang2023near}, i.e., when the learner is allowed to interact with and thereby explore the environment, in theory, little is known about how such policy can be learned with provable efficiency in the offline setting, where the learner has a pre-collected dataset but cannot interact with the environment \citep{urpi2021risk,ma2021conservative,rigter2024one}.

Learning from interactions with the environment can be cost-prohibitive, thereby preventing us from actually learning and benefiting from these policies. Take the financial applications of RL for instance. Supposing we are training an RL agent to optimize the portfolio in the stock market (see~\citet{yu2019model, huang2021novel, chaouki2020deep} for examples of such attempts), training an RL agent from scratch via online interactions in this setting may lead to significant financial losses, as these agents often require long periods of interacting with the environment before recovering a satisfactory policy. As such, many existing works focus on utilizing pre-collected datasets to learn an effective policy, falling into the offline RL setting. Moreover, many practical problems, such as financial applications, are highly risk-sensitive in nature. A rigorous theoretical analysis would help us better understand the possibilities and impossibilities of offline RL methods in such applications. Our work thus concentrates on answering the following critical question:

\vspace{3pt}
{\centering
\emph{Can we design a provably efficient risk-sensitive RL algorithm with an offline dataset?}
\par
}
\vspace{2pt}

Our work takes an initial step toward answering this question. Inspired by a long line of related works investigating risk-sensitivity in sequential decision-making~\citep {howard1972risk,jaskiewicz2007average,bauerle2014more,osogami2012robustness,tamar2012policy,patek2001terminating,shapiro2021lectures,shen2014risk,borkar2001sensitivity,mihatsch2002risk,borkar2002q,borkar2002risk,di2007infinite,ma2020dsac,zhou2023risk,moharrami2024policy}, we investigate optimizing the following entropic risk measure in Markov Decision Processes (MDPs), defined as
\begin{equation}\label{eq:entropic-risk-measure}
    \textstyle
    V_\beta := \frac{1}{\beta}\log\big\{\EE \big[e^{\beta R}\big]\big\},
\end{equation}
with $\beta \in \RR$ being an adjustable parameter controlling the risk-sensitivity. Consider the second-order Taylor expansion of~\eqref{eq:entropic-risk-measure} for intuition, where $V_\beta = \EE[R] +\frac{\beta}{2}\mathrm{Var}[R]+O(\beta^2)$. When $\beta = 0$, we recover the risk-neutral objective. When $\beta < 0$, the objective is risk-averse, and for $\beta$'s further away from 0, the objective further penalizes trajectories with larger reward variances. On the other hand, setting $\beta > 0$ leads to a risk-seeking objective, which can be used to model risk-seeking human behavior in psychology applications~\citep{braun2011risk, chateauneuf1994risk}.
Moreover, to tackle the issue of large state space that can result in increased sample complexity in the tabular setting, our work considers the linear MDP, a popular theoretical framework for function approximation. 

To the best of our knowledge, no existing work studies provably efficient offline RL with respect to the entropic risk measure in the tabular MDP setting, let alone the more general linear function approximation setting. 
Moreover, while it is common practice to design a pessimistic algorithm for risk-neutral offline RL, it remains elusive how pessimism could be implemented for risk-sensitive RL with the special entropic risk measure we focus on. Additionally, we also aim to design a variance-aware algorithm to sharpen our rates. However, we are unaware of any existing works incorporating variance estimation in risk-sensitive RL, even in those that focus on risk-sensitive online RL. Thus, it is technically challenging to provide the theoretical guarantees incorporating the entropic measure in offline RL for our proposed algorithms.
In this work, by tackling the above challenges, we successfully propose two pessimistic algorithms to learn the optimal policy for the offline risk-sensitive RL with provable guarantees under the linear function approximation setting. 
We summarize our main contributions below.


{\noindent \textbf{Contributions.}} In this paper, we propose the first provably efficient risk-sensitive RL algorithms with linear function approximation for the offline setting. Specifically, our first algorithm is a pessimistic value iteration algorithm with a pessimistic bonus term devised using the structure of the entropic measure for eliminating spurious correlation. In addition, utilizing both variance information and reference-advantage decomposition, we develop a variance-aware pessimistic value iteration algorithm by devising a variance estimator for the entropic value function, aiming to improve our theoretical guarantee further. In our theoretical results, we show that the first algorithm is sufficiently efficient to learn an optimal policy with a guarantee depending on a risk-sensitivity factor and feature space dimension $d$, and then it can achieve a tighter guarantee using reference-advantage decomposition with an improved feature space dimension dependence from $d$ to $\sqrt{d}$ with an additional coverage assumption. Moreover, we prove that our variance-aware algorithm can effectively sharpen both the dependence on the feature space dimension $d$ and the risk-sensitivity factor. When we take $\beta \to 0^+$, we can recover the best-known rate for risk-neutral offline RL. 
In our proofs, we provide a novel analysis of the covering number for risk-sensitive value function estimates in linear MDPs, which has not been studied in previous works, even under the online setting, making it of independent interest.

\subsection{Related Work} 
Our work is related to a long line of works using linear function approximation in single-agent RL with an unknown transition and reward function. Particularly, whereas some works focus on the linear mixture MDPs \citep{zhou2021nearly, hu2022nearly, chen2022sample, zhou2021provably,cai2020provably,zhang2021reward,yang2019sample}, where each state, action, next state tuple is mapped into a feature space, we focus on the linear MDP case where features depend on the state and action only \citep{jin2020provably,zanette2021cautiously,jin2021pessimism, wang2020reward,wagenmaker2022first,zhong2024theoretical,liu2024optimistic,agarwal2020pc}. As the lines of research closely relevant to the linear MDP, the recent works further study low-rank MDPs \citep{agarwal2020flambe,zhou2020nonstationary,he2021logarithmic,uehara2021representation,min2021variance,qiu2022contrastive,zhang2022making,zheng2022optimistic,mhammedi2024efficient,modi2024model}, where the state-action feature is learned instead of known a prior, and the kernel function approximation \citep{yang2020provably,qiu2021reward}, which covers the linear MDP as a special case. We leave the study of extending our method for linear MDPs to such settings as our future work.

We also draw inspiration from a line of work on risk-neutral offline RL \citep{levine2020offline} for the single-agent MDP, where the agent aims to recover a near-optimal policy from a pre-collected dataset under an unknown transition and reward function \citep{rashidinejad2021bridging,jin2021pessimism, xie2021bellman, cheng2022adversarially, zanette2021provable,uehara2021pessimistic,xiong2022nearly,yin2022near,shi2022pessimistic,yin2021towards,nguyen2023instance,di2023pessimistic,li2024settling}. Of these works,~\citet{xiong2022nearly,yin2022near} achieve the tightest bounds for linear MDPs that we are aware of, and our work is able to recover their rates under similar assumptions when $\beta \rightarrow 0$ (i.e., risk-neutral). While the algorithm discussed in Section~\ref{sec:va-rspvi} is partially inspired by these works, it is important to emphasize that using variance estimation and reference-advantage decomposition in our risk-sensitive setting requires careful analysis of the specific problem structure defined by the entropic risk measure, which thus is not straightforward given existing works.

Our work is closely related to a line of works studying the entropic risk measure in RL  \citep{jaskiewicz2007average,shen2013risk,bauerle2014more,osogami2012robustness,patek2001terminating,nass2019entropic,shapiro2021lectures,shen2014risk,borkar2001sensitivity,mihatsch2002risk,borkar2002q,borkar2002risk,zhang2023regularized,di2007infinite,ma2020dsac,moharrami2024policy}, dating back to \citet{howard1972risk}, which is consistent with the same risk measure's usage in financial mathematics~\citep{detlefsen2005conditional, rudloff2008entropic}. Particularly, among recent works, \citet{fei2021exponential, fei2021risk, fei2020risk, fei2022cascaded,liang2022bridging} theoretically study online reinforcement learning under the entropic risk measure with theoretical guarantees, whereas we focus on the offline setting. Moreover, only~\citet{fei2021risk} incorporates linear function approximation in the model, and the rest focuses only on tabular MDPs. Even for~\citet{fei2021risk}, the work only focuses on linear mixture MDPs, avoiding a challenging covering number analysis for the entropic risk value function estimates under linear MDPs. Orthogonal to our studies is another line of literature studying risk-sensitive RL with the CVaR risk measure~\citep{prashanth2014policy,lim2022distributional,bastani2022regret,du2023provably, wang2023near,zhao2023provably,chen2023provably} or general risk measures~\citep{wu2023risk} defined by different utility functions \citep{follmer2002convex,ben2007old,lee2020learning}. It is interesting to further extend our analysis of the entropic risk measure to these various risk measures.

\section{Preliminaries}

\textbf{Markov Decision Process.} We study an episodic MDP characterized by a tuple $(\cS,\cA,\PP, r, H)$, where $\cS$ is a possibly infinite state space, $\cA$ the agent's action space, $H$ the episode length, $\PP:=\{\PP_h\}_{h=1}^H$ the transition kernel with $\PP_h(s' | s,a)$ being the probability density of transitioning from the current state $s$ to the next state $s'$ upon taking action $a\in \cA$ at step $h$, and $r=\{r_h\}_{h=1}^H$ the reward function where $r_h: \cS\times\cA \mapsto [0,1]$. 
We define $\pi=\{\pi_h\}_{h=1}^H$ with $\pi_h(\cdot|s)\in \Delta_{\cA}$  for all $s\in \cS $ as the policy for the agent where $\Delta_\cA$ is the set of probability measures on $\cA$. 
For interacting with the environment at step $h$, the agent at a state $s_h$ takes an action $a_h\sim \pi_h(s_h)$ and transitions to the next state $s_{h+1}\sim \PP_h(\cdot\given s_h,a_h)$, receiving a reward $r_h(s_h,a_h)$ in the process. Without loss of generality, we assume that the interaction always starts from a fixed initial state $s_1$.  
We further assume that the transition kernel $\PP$ and the reward function $r$ are \emph{unknown} to better reflect the challenges in real-world problems.

\textbf{Linear MDP.} When facing a large state space that may result in increased sample complexity in a tabular MDP, a common technique is making use of function approximation, such as the linear function approximation. In this paper, we consider the widely-studied linear MDP model, which admits a linear structure in both the reward function and the transition kernel, i.e.,
\begin{align}
\begin{aligned}
\label{eq:lin-mdp}
    &r_h(s,a) = \theta_h^\top \phi(s,a), \\
    &\PP_h(\cdot|s,a) = \mu_h(\cdot)^\top \phi(s,a),  
\end{aligned}
\end{align}
where $\max\{\int_\cS \|\mu_h(s)\| \id s, \|\theta_h\|\}\leq \sqrt{d}$, and $\phi:\cS\times\cA\mapsto \RR^d$ is a feature map with $\|\phi(s,a)\|\leq 1$. In particular, with finite actions and states, by setting $d=|\cS| |\cA|$ and $\phi(s, a) = \be_{(s,a)}$ as the canonical basis, it reduces to the tabular MDP setting where the states and actions are discrete.


 
\textbf{Value Function and Optimal Policy.} 
Recall the entropic risk measure in~\eqref{eq:entropic-risk-measure}. For risk-sensitive RL, we define the state-value function (Q-function) $Q_h^\pi:\cS \times\cA \mapsto \RR$ as the expected cumulative rewards, measured by the entropic risk measure, under some policy $\pi$ starting from a state-action tuple $(s,a)$. More concretely, we let

\vspace{-0.55cm}
\small
\begin{align*}
&Q_h^\pi(s,a)=\frac{1}{\beta} \log\big\{\EE_{\pi}\big[  e^{\beta \sum_{h'=h}^H r_{h'}(s_{h'},a_{h'})} \given s_h=s,a_h=a\big]   \big\},
\end{align*}
\normalsize
\vspace{-0.6cm}

where the expectation is taken over the trajectories induced by $\pi$. The value function $V_h^\pi:\cS \mapsto \RR$ is defined as the expected cumulative rewards under the policy $\pi$ starting at a state $s$, which is 

\vspace{-0.55cm}
\small
\begin{align*}
    V_h^\pi(s)=\frac{1}{\beta} \log\big\{\EE_\pi\big[  e^{\beta \sum_{h'=h}^H r_{h'}(s_{h'},a_{h'})}~\biggiven~s_h=s \big]\big\}.
\end{align*}
\normalsize
\vspace{-0.6cm}

We introduce the following shorthand notations for the conditional expectation and variance of any function $f:\cS\to\RR$, taken over the randomness in the next step transition
\begin{align*}
    \textstyle
    &(\PP_h f)(s,a) = \EE_{s'\sim \PP_h(\cdot|s,a)}[f(s')|s_h=s,a_h=a], \\
    &(\mathrm{Var}_h f)(s,a) =  (\PP_h f^2)(s,a) -  [(\PP_h f)(s,a) ]^2.
\end{align*}
Letting $V_{H+1}^\pi(s) = Q_{H+1}^\pi(s,a)=0$ for any $(s,a)$, we have the following Bellman equation,
\begin{align}
\begin{aligned}\label{eq:bellman}
    \textstyle
    &Q_h^\pi(s,a) = r_h(s,a) + \frac{1}{\beta}\log\left((\PP_h e^{\beta V_{h+1}^\pi})(s,a)\right), \\
    &V_h^\pi(s) = \frac{1}{\beta}\log\left(\inner{e^{\beta Q_h^\pi(s,\cdot)},\pi(\cdot|s)}_\cA\right). 
\end{aligned}
\end{align}
where $\inner{\cdot,\cdot}_\cA$ is the inner product over $\cA$. The policy $\pi$ can be stochastic in our setting. Based on the definition of the value function, we define the optimal policy $\pi^*$ as $\pi^*:=\argmax_{\pi} V_1^{\pi}(s_1)$. For brevity, we use $Q_h^*$ and $V_h^*$ to denote the optimal Q-function and optimal value function, respectively. Their relationship is characterized by the exponential Bellman optimality equation
\begin{align*}
    &e^{\beta Q_h^*(s,a)} = e^{\beta r_h(s,a)}\cdot(\PP_h e^{\beta V_{h+1}^*})(s,a),\\
    &V_h^*(s) =\max_{a\in\cA}Q_h^*(s,a).
\end{align*}
Our goal is to learn a policy $\pi$ that maximizes $V_1^{\pi}(s_1)$, and we measure the policy's performance by 
\begin{align*}
 \mathrm{SubOpt}(\pi):= V_1^{*}(s_1) - V_1^{\pi}(s_1),   
\end{align*}
that is, the suboptimality of the policy $\pi$ given that the initial state is $s_1$. In addition, we say a policy $\pi$ is $\varepsilon$-approximate optimal if $\mathrm{SubOpt}(\pi) \leq \varepsilon$.

\textbf{Risk-Sensitive Offline RL.} In offline RL, we assume that there exists a behavior policy that generates an offline dataset by interacting with the environment. With this dataset, we then employ offline RL algorithms to recover an estimate of $\pi^*$ without interacting with the environment. Particularly, we make the following assumptions about the behavior policy and the offline dataset:
\begin{assumption}[Offline Dataset] \label{assump:offline-data} Suppose that there exists a behavior policy $\mu$. After interacting with the environment for $K$ rounds, we sample a dataset $\cD$ consisting of $K$ trajectories, which is defined as $\cD:=\{(s_h^\tau, a_h^\tau, r_h(s_h^\tau, a_h^\tau))\}_{h,\tau=1}^{H,K}$.
\end{assumption}
Assumption~\ref{assump:offline-data} does not enforce any coverage conditions on the dataset but precludes us from using techniques such as variance estimation and reference-advantage decomposition. As such, we make the following stronger but still common assumption on data coverage~\citep{xiong2022nearly,wang2020statistical,duan2020minimax,yin2022near,yin2021towards}, with which we can obtain the improved dependence on $d$ and tighter suboptimality guarantees.
\begin{assumption}[Data Coverage] \label{assump:coverage} We assume that the smallest eigenvalue of the covariance matrix is bounded away from zero, that is
\begin{align*}
\kappa = \min_{h\in[H]}\lambda_{\min}(\EE_{d_h^\mu}[\phi(s,a)\phi(s,a)^\top])>0,   
\end{align*}
where $d_h^\mu$ denotes the joint distribution of $(s,a)$ over $\cS \times \cA$ at step $h$ induced by the behavioral policy $\mu$.
\end{assumption}
We note that Assumption~\ref{assump:coverage} is not essential, and our algorithm can still efficiently learn a near-optimal policy without it, albeit at a slower rate. In the section on main results, we first prove a result for learning an $\varepsilon$-approximate optimal policy without this assumption in Theorem \ref{thm1}. Then, we target to obtain sharper guarantees based on Assumption \ref{assump:coverage}. These sharper results are further provided in Theorem \ref{thm2} using the same algorithm for Theorem \ref{thm1}, and in Theorem \ref{thm3} with designing a different variance-aware algorithm.

\section{Suboptimality Analysis of Risk-Sensitive RL}

To motivate the algorithm design, we provide a high-level overview of the concentration terms we target during analysis and formally introduce the ``shifting and scaling'' technique. We begin by introducing the model evaluation error under our setting, then show how the error relates to the suboptimality of a pessimistic value function estimate, and finally introduce the ``shifting and scaling'' technique, which ensures that the regression targets are on the same scale. 

The suboptimality is closely related to the model evaluation error in offline RL \citep{jin2021pessimism}. However, the standard analysis does not apply in our risk-sensitive setting because the Bellman equation has a different structure, as shown in \eqref{eq:bellman}, necessitating a new definition of model evaluation error. For any estimated optimal Q-function $\wh Q_h$ and value function $\wh V_h$, it seems natural to measure the estimation error of \eqref{eq:bellman} by the difference between $\wh Q_h(s,a)$ and $ r_h(s,a) + \frac{1}{\beta}\log((\PP_h e^{\beta \wh V_{h+1}})(s,a))$. However, the direct approach has been shown to incur an extra factor $e^{\abs{\beta}H^2}$ in the resulting upper bound, even in the tabular MDP setting considered by \citet{fei2021exponential}. Inspired by the work, we instead focus on the \emph{exponential Bellman equation}, obtained by exponentiating both sides of \eqref{eq:bellman},
\begin{align}
\begin{aligned}\label{eq:ebellman}
    &e^{\beta Q_h^\pi(s,a)} =e^{\beta r_h(s,a)}(\PP_h e^{\beta V_{h+1}^\pi})(s,a),\\
    &e^{\beta V_h^\pi(s)} =\inner{e^{\beta Q_h^\pi(s,\cdot)},\pi(\cdot|s)}_\cA. 
\end{aligned}
\end{align}
The model evaluation error is then defined as
\begin{align*}
    \iota_{\mathrm{exp},h}(s,a) = e^{\beta r_h(s,a)}\PP_h e^{\beta\wV_{h+1}}(s_h,a_h)-e^{\beta\wQ_h(s,a)},
\end{align*}
where the subscript $\exp$ denotes that the evaluation error is defined with respect to the exponential Bellman equation.

Following the spirit of the exponential Bellman equation in \eqref{eq:ebellman}, we leverage model evaluation error and pessimism to control $e^{\beta V_h^*(s)} - e^{\beta V_h^{\wh\pi}(s)}$ directly, as opposed to controlling the term via bounding $V_h^*(s) - V_h^{\wh\pi}(s)$.
\begin{lemma}
    \label{lemma:pessimistic-bound-of-suboptimality}
    Let $\wh{Q}$ be a pessimistic estimate of $Q$, satisfying $\mathrm{sign}(\beta)\iota_{\mathrm{exp}, h}(s, a) \geq 0$ for any $(s,a,h)\in \cS\times\cA\times [H]$. $\wh{V}_{h}(s) = \max_{a \in \cA}\wh{Q}_{h}(s, a)$ denote the value function induced by $\wh{Q}$. Then, we have for all $\beta > 0$,
    \begin{align*}
    \textstyle
        \mathrm{SubOpt}(\wh\pi)&\le\frac{1}{\beta}\left(e^{\beta V_1^{*}(s_1)}-e^{\beta V_1^{\wh\pi}(s_1)}\right)\\
        & \leq \sum_{h=1}^H\frac{e^{\beta (h-1)}}{\beta}\EE_{\pi^*}\left[\iota_{\mathrm{exp},h}(s_h,a_h)|s_1\right],
    \end{align*}
     and for all $\beta < 0$,
    \begin{align*}
    \textstyle
         \mathrm{SubOpt}(\wh\pi)&\le\frac{e^{-\beta H}}{\beta}\left(e^{\beta V_1^{*}(s_1)}-e^{\beta V_1^{\wh\pi}(s_1)}\right) \\
         &\leq \frac{e^{-\beta H}}{\beta}\sum_{h=1}^H\EE_{\pi^*}\left[\iota_{\mathrm{exp},h}(s_h,a_h)|s_1\right],
    \end{align*}
    where $\wh{\pi}$ is the greedy policy taken with respect to $\wh{Q}$, defined by $\wh{\pi}_{h}(s) = \argmax_{a \in \cA} \wh{Q}_{h}(s, a)$.
\end{lemma}
We can observe that the model evaluation errors propagate backward differently. When $\beta < 0$ (risk-averse), the model valuation errors at each step scale the same in the bound, whereas errors at later steps are scaled upward when $\beta > 0$ (risk-seeking). The difference in scale, while inevitable, can be better accounted for via the following ``shifting and scaling'' technique.

\textbf{Shifting and Scaling.} We define the following shifting and scaling transformation:
\begin{align}\label{scaling}
    \SSS_h f(\cdot) =
    \begin{cases}
     e^{\beta (h-1)}( e^{\beta f(\cdot)}-1),\quad &\beta>0\\
     -e^{-\beta H}(e^{\beta f(\cdot)} - 1),\quad &\beta <0
    \end{cases}
\end{align}
for any function $f$ satisfying $f(s)\in[0,H+1-h]$.  With the operator defined, at each step, rather than regressing directly on $\exp(V_h(s'))$, with $s'$ denoting the observed state at the next step, we instead regress on $\SSS_{h}V_{h}(s')$ to rescale the regression targets at each step.

For the case $\beta > 0$, there is a scaling factor of $e^{\beta(h - 1)}$ in \eqref{scaling}, ensuring that the regression targets at each step are scaled according to the decomposition bounds in Lemma~\ref{lemma:pessimistic-bound-of-suboptimality}. For $\beta < 0$, the scaling factor $e^{-\beta H}$  is similarly constructed to ensure that the regression targets are on the same scale as the suboptimality bounds. Together with the shifting term, the range of the value function after the transformation satisfies $\SSS_hV_h(\cdot) \in [0,e^{\abs{\beta}H}-1]$ for any $\beta\neq 0$ and $h$. While shifting does not affect the scale of the model evaluation error, it ensures that the error's range starts at zero. For concentration analysis, we can then focus only on regression targets in the range $[0,e^{\abs{\beta}H}-1]$, regardless of the sign of $\beta$ or the value of $h$. Our shifting and scaling operator integrates principles from shifted exponential V-functions in \cite{fei2021risk} and decaying bonus in \cite{fei2021exponential} from a unified perspective. Additionally, our operator can rescale the target in advance and thus avoid adjusting the uncertainty bonus at each step as in \cite{fei2021exponential}.

This method provides multiple benefits. To understand the benefits of shifting, we examine our resulting upper bound when $\beta>0$, which depends on $\beta$ through a risk-sensitivity factor, $\frac{e^{\beta H}-1}{\beta}$. If we overlook the property that $e^{\beta V_h}\ge 1$ and merely treat $e^{\beta V_h}$ as a function bounded by $e^{\beta H}$ without shifting it, the term $e^{\beta H}-1$ in the upper bound will be replaced with $e^{\beta H}$. By shifting $e^{\beta V_h}$ by 1, the upper bound could be improved by a factor $(e^{\beta H}-1)/e^{\beta H}< 1$, which could be much smaller, particularly when $\beta$ approaches zero. Additionally, scaling is also beneficial. As shown in \ref{lemma:pessimistic-bound-of-suboptimality}, the errors are scaled upward when $\beta>0$. If errors at different steps are treated uniformly without incorporating the fact that $V_h$s have different ranges, the additional $e^{|\beta|H^2}$ term in the resulting upper bound in \cite{fei2021risk} cannot be entirely eliminated. Moreover, as the regression targets are scaled and shifted correctly, we avoid the need to derive suitable concentration bounds over the different ranges of regression targets caused by the choices of $\beta$ and $h$, streamlining the analysis. 

%

\section{Algorithm} \label{sec:alg}

In this section, we propose two offline algorithms for risk-sensitive RL under the linear MDP setting. 

\begin{algorithm}[t!]
\caption{RSPVI Algorithm}
\label{alg:alg1}
\setstretch{1.1}
\begin{algorithmic}[1]
\STATE \textbf{Input:} Dataset $\cD:=\{(s_h^\tau, a_h^\tau, r_h(s_h^\tau, a_h^\tau))\}_{h,\tau=1}^{H,K}$.
\STATE \textbf{Initialize:} $\widehat V_{H+1}(\cdot)=0$.
\FOR{$h=H,\dots,1$}
    \STATE $\Lambda_h \gets \sum_{\tau=1}^K\phi(s_h^\tau,a_h^\tau)\phi(s_h^\tau,a_h^\tau)^\top +\lambda I_d$
    \STATE $\wh\theta_h \gets \invL\sum_{\tau=1}^K\phi(s_h^\tau,a_h^\tau)r_h(s_h^\tau,a_h^\tau)$
    \STATE $\widehat w_h \gets \Lambda_h^{-1}\sum_{\tau=1}^K\phi(s_h^\tau,a_h^\tau) \SSS_{h+1} \wh V_{h+1}(s_{h+1}^\tau)$
    \STATE $\wh r_h(\cdot,\cdot)\gets\big\{\phi(\cdot,\cdot)^\top \wh\theta_h\big\}_{[0,1]}$
    \STATE $\Gamma_h(\cdot,\cdot) \gets \gamma\Lnorm{\phi(\cdot,\cdot)}$    
    \STATE $\widehat Q_h(\cdot,\cdot) \gets \big\{\frac{1}{\beta}\log(q_h(\cdot,\cdot))\big\}_{[0,H+1-h]}$ where $q_h$ is 
    \begin{itemize}[leftmargin=*,topsep=5pt,itemsep=-8pt,parsep=0pt,labelindent=5mm]
        \item \hspace{-0.39cm}when $\beta>0$,
            \vspace{-0.15cm}
            \begin{align*}
                \hspace{-0.7cm}e^{\beta (1-h)}\big[e^{\beta(\widehat r_h(\cdot,\cdot)-1)}(\phi(\cdot,\cdot)^\top \widehat w_h +e^{\beta h})-\Gamma_h(\cdot,\cdot)\big] 
            \end{align*}
        \item \hspace{-0.39cm}when $\beta<0$,
            \vspace{-0.15cm}
            \begin{align*}
                \hspace{-1.46cm}e^{\beta H}\big[e^{\beta\widehat r_h(\cdot,\cdot)}(e^{-\beta H}-\phi(\cdot,\cdot)^\top \widehat w_h) +\Gamma_h(\cdot,\cdot)\big]
            \end{align*}
    \end{itemize}    
    \vspace{-0.2cm}
    \STATE $\widehat\pi_h(\cdot|\cdot) \gets \argmax_{\pi_h} \langle \widehat Q_h(\cdot,\cdot),\pi_h(\cdot|\cdot)\rangle_\cA$
    \STATE $\widehat V_h(\cdot) \gets\frac{1}{\beta}\log\big(\langle \exp(\beta\widehat Q_h(\cdot,\cdot)),\widehat\pi_h(\cdot|\cdot)\rangle_\cA\big)$
\ENDFOR
\STATE \textbf{Output: $\widehat\pi =\{\widehat\pi_h\}_{h=1}^H$}
\end{algorithmic}
\end{algorithm}

\subsection{Risk-Sensitive Pessimistic Value Iteration}

We begin by introducing the algorithm Risk-Sensitive Pessimistic Value Iteration (RSPVI) as summarized in Algorithm \ref{alg:alg1}. The algorithm's performance relies on the shifting and scaling technique, and its construction draws inspiration from \citet{jin2021pessimism}.

Specifically, Lines 4-6 in Algorithm \ref{alg:alg1} perform two ridge regressions, and $\wh\theta_h$ and $\wh w_h$ are the solutions of the following minimization problems:
\begin{align*}
   \min_{\theta\in\mathbb R^d}~\sum_{\tau = 1}^K \Big[r_h(s_h^\tau,a_h^\tau)-\phi_h(s_h^\tau,a_h^\tau)^\top \theta\Big]^2 +\lambda\norm{\theta}_2^2, 
\end{align*}
and
\begin{align*}
    \min_{w\in\mathbb R^d}~&\sum_{\tau = 1}^K \Big[\SSS_{h+1} \wh V_{h+1}(s_{h+1}^\tau)-\phi_h(s_h^\tau,a_h^\tau)^\top w\Big]^2 + \lambda\norm{w}_2^2.
\end{align*}
Combining such results, the estimated reward functions are constructed by $\wh r_h(\cdot,\cdot)=\big\{\phi(\cdot,\cdot)^\top \wh\theta_h\big\}_{[0,1]}$, where $\{z\}_{[x,y]}$ clips $z$ into the range $[x, y]$, i.e., $\{z\}_{[x, y]} = \max\{x,\min\{z,y\}\}$. Our approach is justified in the linear MDP setting, as the transition operator $\PP_h$ is linear in features. Therefore $\phi(s,a)^\top \wh w_h$ can be viewed as the estimate of $\PP_h(\SSS_{h+1}\wh V_{h+1})(s,a)$.

In Line 9, we undo the affine transformation $\SSS_h$ and combine the estimates of $r_h(s,a)$ and $\PP_h(\SSS_{h+1}\wh V_{h+1})(s,a)$ into the estimate of $e^{\beta  Q_h}$, or equivalently $e^{\beta r_h}\PP_h (e^{\beta V_{h+1}})$. An uncertainty bonus of $\gamma\Lnorm{\phi(s, a)}$ is subtracted for each $(s, a)$ to enforce pessimism. The pessimistic estimates for $Q, \pi,$ and $V$ can then be obtained following Lines 9-11.

Algorithm~\ref{alg:alg1} is attractive as
it provably recovers a near-optimal policy even under insufficient coverage, achieving a performance guarantee in the absence of Assumption~\ref{assump:coverage}. Moreover, as we show in the sequel, under such a slightly stronger assumption on data coverage (Assumption~\ref{assump:coverage}), the algorithm can achieve a tighter dependence on $d$ with changes to the theoretical analysis only.

\begin{algorithm}[t!]
\caption{VA-RSPVI Algorithm}
\label{alg:alg2}
\setstretch{1.1}
\begin{algorithmic}[1]
\STATE \textbf{Input:} Dataset $\cD:=\{(s_h^\tau, a_h^\tau, r_h(s_h^\tau, a_h^\tau))\}_{h,\tau=1}^{H,K}$ and auxiliary dataset ${\cD}^{\rm aux}:=\{(\breve{s}_h^\tau, \breve{a}_h^\tau, r_h(\breve{s}_h^\tau, \breve{a}_h^\tau))\}_{h,\tau=1}^{H,K}$ 
\STATE \textbf{Initialize:} Set $\widehat V_{H+1}(\cdot)=0$
\STATE Construct variance estimator $\widehat\sigma_h^2(\cdot,\cdot)$ with $\cD^{\rm aux}$ via \eqref{eq:variance-estimator}. Let $(\widehat\sigma_h^\tau)^2 := \widehat\sigma_h^2(s_h^\tau,a_h^\tau)$.
\FOR{$h=H,\dots,1$}
    \STATE  $\Sigma_h \gets \sum_{\tau=1}^K\phi(s_h^\tau,a_h^\tau)\phi(s_h^\tau,a_h^\tau)^\top/(\widehat\sigma_h^\tau)^2 +\lambda I_d$
    \STATE $\wh\theta_h \gets \invS\sum_{\tau=1}^K\phi(s_h^\tau,a_h^\tau)r_h(s_h^\tau,a_h^\tau)/(\widehat\sigma_h^\tau)^2$
    \STATE  $\widehat w_h \gets \Sigma_h^{-1}\sum_{\tau=1}^K\phi(s_h^\tau,a_h^\tau) \SSS_{h+1} \wh V_{h+1}(s_{h+1}^\tau)/(\widehat\sigma_h^\tau)^2$
    \STATE $\wh r_h(\cdot,\cdot) \gets \big\{\phi(\cdot,\cdot)^\top \wh\theta_h\big\}_{[0,1]}$
    \STATE $\Gamma_h(\cdot,\cdot) \gets \gamma\Snorm{\phi(\cdot,\cdot)}$    
    \STATE  $\widehat Q_h(\cdot,\cdot) \gets \big\{\frac{1}{\beta}\log(q_h(\cdot,\cdot))\big\}_{[0,H+1-h]}$ where $q_h$ is 
    \begin{itemize}[leftmargin=*,topsep=5pt,itemsep=-8pt,parsep=0pt,labelindent=5mm]
        \item \hspace{-0.39cm}when $\beta>0$,
            \vspace{-0.15cm}
            \begin{align*}
                \hspace{-0.7cm}e^{\beta (1-h)}\big[e^{\beta(\widehat r_h(\cdot,\cdot)-1)}(\phi(\cdot,\cdot)^\top \widehat w_h +e^{\beta h})-\Gamma_h(\cdot,\cdot)\big]
            \end{align*}
        \item \hspace{-0.39cm}when $\beta<0$,
            \vspace{-0.15cm}
            \begin{align*}
                \hspace{-1.37cm}e^{\beta H}\big[e^{\beta\widehat r_h(\cdot,\cdot)}(e^{-\beta H}-\phi(\cdot,\cdot)^\top \widehat w_h) +\Gamma_h(\cdot,\cdot)\big]
            \end{align*}
    \end{itemize}    
    \vspace{-0.2cm}
    \STATE  $\widehat\pi_h(\cdot|\cdot) \gets \argmax_{\pi_h} \langle \widehat Q_h(\cdot,\cdot),\pi_h(\cdot|\cdot)\rangle_\cA$
    \STATE  $\widehat V_h(\cdot) \gets \frac{1}{\beta}\log\big(\langle \exp(\beta\widehat Q_h(\cdot,\cdot)),\widehat\pi_h(\cdot|\cdot)\rangle_\cA \big)$
\ENDFOR
\STATE \textbf{Output:} $\widehat\pi =\{\widehat\pi_h\}_{h=1}^H$
\end{algorithmic}
\end{algorithm}
 \setlength{\textfloatsep}{20pt}

\vspace{-0.05cm}
\subsection{Variance-Aware RSPVI}\label{sec:va-rspvi}

We further sharpen the suboptimality bound by incorporating variance information and propose Variance-Aware Risk-Sensitive Pessimistic Value Iteration (VA-RSPVI) in Algorithm \ref{alg:alg2}. Particularly, by estimating the variances of the value function estimates, we can reweigh the Bellman residuals to achieve a tighter bound. At a high level, the algorithm first uses Algorithm~\ref{alg:alg1} as a subroutine to obtain variance estimates. A weighted ridge regression is then solved, and the Q-function, policy, and V-function estimates are obtained by solving the weighted regression problem rather than the unweighted ones used by Lines 4-6 of Algorithm~\ref{alg:alg1}. While the technique of utilizing the variance information has been studied in the risk-neural setting \citep{zhou2021nearly,xiong2022nearly,yin2022near}, our work provides a non-trivial generalization of this technique to the risk-sensitive setting due to the exponential Bellman equation.

{We first highlight a nuanced difference between VA-RSPVI and RSPVI. Note that we must ensure that the variance estimate is independent of the regression targets when estimating $\SSS_{h}\hat{V}_{h}$. Otherwise, the variance estimates used to weigh the Bellman errors are correlated with the Bellman residual errors, complicating later convergence analysis. We assume the existence of auxiliary data as a ``reference dataset'', which is sampled from the same distribution as $\cD$ yet independent of the dataset itself \citep{xie2021policy,xiong2022nearly,zhang2022corruption}.}

\begin{assumption}[Auxiliary Offline Dataset] \label{assump:auxiliary-data} Suppose that we have another offline dataset sampled independently following the behavior policy $\mu$ as in Assumption \ref{assump:offline-data}. This auxiliary offline dataset ${\cD}^{\rm aux}$ consisting of $K$ trajectories is defined as ${\cD}^{\rm aux}:=\{(\breve{s}_h^\tau, \breve{a}_h^\tau, r_h(\breve{s}_h^\tau, \breve{a}_h^\tau))\}_{h,\tau=1}^{H,K}$.
\end{assumption}

We make the assumption only for ease of presentation. We note that when no such $\cD^{\rm aux}$ is available, the learner can simply take the original dataset $\cD$ and randomly split it by half over the trajectories, similar to the procedures used by~\citet{xie2021policy, zhang2022corruption}. This will relax our suboptimality bound by only a negligible numerical constant factor and will not change the upper bound's dependence on dominating terms. 

\textbf{Variance Estimation.} By the linear MDP assumption, we know that $[\PP_h (\SSS_{h+1} V^*_{h+1})](s,a) = \phi(s,a)^\top\eta_{h}^{(1)}$ and $[\PP_h (\SSS_{h+1} V^*_{h+1})^2](s,a) =\phi(s,a)^\top\eta_{h}^{(2)}$  for some $\eta_{h}^{(1)}$ and $\eta_{h}^{(2)}$. In other words, the second moment of the value function of the next step, with the expectation taken over the transition dynamics, remains linear in the feature vector $\phi(s, a)$. Using the auxiliary dataset ${\cD}^{\rm aux}:=\{(\breve{s}_h^\tau, \breve{a}_h^\tau, r_h(\breve{s}_h^\tau, \breve{a}_h^\tau))\}_{h,\tau=1}^{H,K}$, we can then calculate the estimate of the weights corresponding to the first moment, denoted as $\wh\eta_h^{(1)}$, as the solution to the following minimization problem
\begin{align*}
 \min_{\eta} \sum_{\tau=1}^K\Big[\phi(\breve{s}_h^\tau,\breve{a}_h^\tau)^\top\eta-\SSS_{h+1}\wV_{h+1}^{\rm aux}(\breve{s}_{h+1}^\tau)\Big]^2+\lambda\norm{\eta}^2_2, 
\end{align*}
and calculate the estimate of the weights corresponding to the second moment, denoted as $\wh\eta_h^{(2)}$, as the solution to another minimization problem
\begin{align*}
\min_{\eta} \sum_{\tau=1}^K\Big[\phi(\breve{s}_h^\tau,\breve{a}_h^\tau)^\top\eta- (\SSS_{h+1}\wV_{h+1}^{\rm aux}(\breve{s}_{h+1}^\tau))^2\Big]^2 +\lambda\norm{\eta}^2_2,  
\end{align*}
where $\hat{V}^{\rm aux}_{h+1}$ is an estimator of $V_{h+1}^*$, obtained by calling Algorithm~\ref{alg:alg1} as a subroutine, with $\cD^{\rm ref}$ as the input. {Doing so thus ensures that the weights $\wh\eta_h^{(1)}$ and $\wh\eta_h^{(2)}$ are independent of the Bellman residual errors calculated using $\cD$, avoiding the potential dependence that may be introduced by variance estimation.} Equipped with $\wh\eta_h^{(1)}$ and $\wh\eta_h^{(2)}$, the conditional variance of $\SSS_{h+1}V_{h+1}^*$ could then be estimated by
\begin{equation}
  \label{eq:variance-estimator}
  \begin{aligned}
    \wh\sigma_h^2(\cdot,\cdot) &:=\max \big\{\underline{\sigma}^2,\{\phi(\cdot,\cdot)^\top \wh\eta_h^{(2)}\}_{[0, (e^{\abs{\beta}H} -1)^2]}\\
    &\qquad \qquad  -\big(\{\phi(\cdot,\cdot)^\top \wh\eta_h^{(1)}\}_{[0, e^{\abs{\beta}H} -1]}\big)^2 \big\}.
  \end{aligned}
\end{equation}
We clip the estimator from below by $\underline{\sigma}^2$ to avoid variance close to 0 and $\wh\sigma_h^2$ is a consistent estimate of the clipped conditional variance $\sigma_h^2:= \max\{\underline{\sigma}^2, \mathrm{Var}_h(\SSS_{h+1}V_{h+1}^*)(s,a)\}$ under the conditions in Theorem \ref{thm3}.

With the variance estimators $\wh\sigma_h^2(s,a)$ on hand, we develop Algorithm \ref{alg:alg2}, namely Variance-Aware Risk-Sensitive Pessimistic Value Iteration, where the coefficients are constructed by weighted ridge regressions. 
The use of $\cD^{\rm aux}$ in Line 3 of Algorithm \ref{alg:alg2} guarantees the independence between the variance estimate and the regression targets. Weighted by the variance estimators, $\wh w_h$ and $\wh\theta_h$ are the solutions of 
\begin{align*}
    \min_{\theta\in\mathbb R^d}~\sum_{\tau = 1}^K \frac{(r_h(s_h^\tau,a_h^\tau)-\phi_h(s_h^\tau,a_h^\tau)^\top \theta)^2}{\widehat\sigma_h^2(s_h^\tau,a_h^\tau)} +\lambda\norm{\theta}_2^2,
\end{align*}
and
\begin{align*}
    \min_{w\in\mathbb R^d}~&\sum_{\tau = 1}^K \frac{(\SSS_{h+1} \wh V_{h+1}(s_{h+1}^\tau)-\phi_h(s_h^\tau,a_h^\tau)^\top w)^2}{\widehat\sigma_h^2(s_h^\tau,a_h^\tau)}+ \lambda\norm{w}_2^2.
\end{align*}
Their closed forms are given by Lines 5-7 of Algorithm \ref{alg:alg2}. The bonus function on Line 9 of Algorithm~\ref{alg:alg2} changes correspondingly, and the Q-function, policy, and value function estimates are obtained accordingly on Lines 10-12. 

Intuitively, by weighing each observation according to their residuals' variance, the procedure is now akin to generalized least squares with $l_2$ regularization~\citep{amemiya1985advanced}. As we prove in the sequel, as long as the sample size is sufficiently large, the estimates' quality in Algorithm~\ref{alg:alg2} will never be worse than that of Algorithm~\ref{alg:alg1} under Assumption~\ref{assump:coverage}, thereby further improving our suboptimality bounds.


\section{Main Results}

In this section, we present our main theoretical results for both Algorithm~\ref{alg:alg1} and Algorithm \ref{alg:alg2}.

{\noindent \bf Suboptimality Bound for Algorithm \ref{alg:alg1}.}
We begin with our baseline result characterizing the suboptimality of Algorithm \ref{alg:alg1}. The result does not require any coverage assumptions and can be sharpened by the utilization of reference-advantage decomposition, as we will discuss later. 
\begin{theorem}\label{thm1}
Under Assumption \ref{assump:offline-data}, if we set $\lambda = 1/e^{2\abs{\beta}}$ and $\gamma = \tilde O(d(e^{\abs{\beta}H}-1))$ in Algorithm \ref{alg:alg1}, with probability at least $1-\delta$, the suboptimality $\mathrm{SubOpt}(\wh\pi):= V_1^*(s_1) - V_1^{\wh\pi}(s_1)$ admits an upper bound of
\begin{align*}
\tilde O\left(d\right)\frac{e^{\abs{\beta}H}-1}{\abs{\beta}}\sum_{h=1}^H \EE_{\pi^*}\left[\Lnorm{\phi(s_h,a_h)}\Big|s_1\right].
\end{align*}
\end{theorem}
Here $\tilde O(\cdot)$ omits terms that are logarithmic in $d$, $H$, $K$, and $1/\delta$, and the proof is deferred to the appendix. As the setting of risk-sensitive tends to the risk-neutral setting when $\beta\to0^+$, one may achieve an upper bound $\tilde O(dH)\sum_{h=1}^H[\Lnorm{\phi(s,a)}|s_1]$, which coincides with the bound in \citet{jin2021pessimism} up to logarithmic factors. 

We stress that Theorem~\ref{thm1} does not require Assumption~\ref{assump:coverage}. We only need to assume the existence of an offline dataset and make no coverage assumption on the dataset. The coverage assumption, Assumption~\ref{assump:coverage}, can further improve our dependence on the feature dimension $d$ by changing the strength of the uncertainty bonus $\gamma$, without any changes to the Algorithm~\ref{alg:alg1}. We present our result as follows:
\begin{theorem}\label{thm2}
Under Assumptions \ref{assump:offline-data}-\ref{assump:coverage}, with $K\ge \tilde\Omega\left(d^2H^2/\kappa +1/\kappa^2\right)$, if we set $\lambda = 1/e^{2\abs{\beta}}$ and $\gamma = \tilde O(\sqrt{d}(e^{\abs{\beta}H}-1))$ in Algorithm \ref{alg:alg1}, with probability at least $1-\delta$,the suboptimality $\mathrm{SubOpt}(\wh\pi):= V_1^*(s_1) - V_1^{\wh\pi}(s_1)$ admits an upper bound of
\begin{align*}
    \tilde O\big(\sqrt{d}\big)\frac{e^{\abs{\beta}H}-1}{\abs{\beta}}\sum_{h=1}^H \EE_{\pi^*}\left[\Lnorm{\phi(s_h,a_h)}\Big|s_1 \right].
\end{align*}
\end{theorem}
Leveraging Assumptions~\ref{assump:coverage}, we show in Theorem~\ref{thm2} that we can improve the dependence on the feature dimension from $\Tilde{O}(d)$ to $\tilde{O}(\sqrt{d})$ by only changing the hyperparameter choice and the analysis of the algorithm. As we detail in Section~\ref{sec:theoretical-analysis}, this improvement is due to the reference-advantage decomposition, which, to the best of our knowledge, has not been applied in risk-sensitive RL by existing works.

{\noindent \bf Suboptimality Bound for Algorithm \ref{alg:alg2}.}
Theorem~\ref{thm3} provides the performance bound for Algorithm~\ref{alg:alg2}. Our theoretical analysis is associated with a term $\xi(\underline{\sigma}^{2}) :=\sup_{h,s,a,s'\sim\PP_h(\cdot|s,a)}\frac{(\SSS_{h+1}V^*_{h+1}(s') - \PP_h (\SSS_{h+1}V^*_{h+1})(s,a))^2}{\max\{\underline{\sigma}^2, \mathrm{Var}_h(\SSS_{h+1}V_{h+1}^*)(s,a)\}}$ that characterizes the degree of deviation from the mean of $\SSS_{h+1}V^*_{h+1}(s')$ standardized by the truncated variance. In Theorem \ref{thm3}, we assume $\xi(\underline{\sigma}^{2}) = O(d)$. We note that \citet{yin2022near} imposed a similar condition with $\underline\sigma^2$ fixed at 1, while we allow the flexibility of adjusting $\underline\sigma^2$.

\begin{theorem}\label{thm3}
Under Assumption \ref{assump:offline-data}-\ref{assump:coverage}, if we have 
$
     K\ge \tilde\Omega(d^2H^2/\kappa +1/\kappa^2)\cdot ((e^{\abs{\beta}H}-1)/\underline\sigma)^4
$
and $\xi(\underline{\sigma}^{2}) = O(d)$,
setting $\lambda  = 1/(e^{\abs{\beta}(H+1)}-e^{\abs{\beta}})^2$ and $\gamma = \tilde O(\sqrt{d})$ in Algorithm \ref{alg:alg2}, with probability at least $1-\delta$, $\mathrm{SubOpt}(\wh\pi):= V_1^*(s_1) - V_1^{\wh\pi}(s_1)$ admits an upper bound of
\begin{align*}
    \tilde O\big(\sqrt{d}\big)\frac{1}{\abs{\beta}}\sum_{h=1}^H \EE_{\pi^*}\left[\norm{\phi(s_h,a_h)}_{(\Sigma_h^*)^{-1}}\Big|s_1 \right],
\end{align*}
where $\Sigma_h^* = \sum_{\tau=1}^K\phi(s_h^\tau,a_h^\tau)\phi(s_h^\tau,a_h^\tau)^\top/\sigma_h^2(s_h^\tau,a_h^\tau) +\lambda I_d$ and $\sigma_h^2(s_h,a_h)$ is the clipped conditional variance $\max\{\underline{\sigma}^2, \mathrm{Var}_h(\SSS_{h+1}V_{h+1}^*)(s,a)\}$.
\end{theorem}
The condition $\xi( \underline{\sigma}^{2} ) = O(d)$ can be ensured by using a sufficiently large $\underline{\sigma}^2$, which in turn guarantees that the magnitude of $\SSS_{h+1} V^*_{h+1}$ on the model evaluation error is negligible. A feasible choice is to set $\underline\sigma^2 \ge (e^{\abs{\beta}H}-1)^2/d$, which ensures $\xi(\underline{\sigma^2}) = O(d)$. One extreme case is $\underline\sigma^2 = (e^{\abs{\beta}H}-1)^2$. In that case, $\xi(\underline{\sigma}^2)=O(1)$ and $\wh\sigma^2_h(s,a)=\underline\sigma^2$ for any $h,s,a$. In other words, in this extreme case, all targets in the regression are given the same weights and $\Sigma_h^*=\Lambda_h/\underline{\sigma}^2$. Consequently, Algorithm \ref{alg:alg2} is equivalent to Algorithm \ref{alg:alg1}, and Theorem \ref{thm3} provides the same upper bound as the one in Theorem \ref{thm2} with appropriate choices of $\gamma$ and $\lambda$.

Our bound implicitly depends on $e^{|\beta|H}$ through the scaled covariance matrix $\Sigma_h^*$ and $\lambda$, as now $\lambda^{-1} = e^{2|\beta|}(e^{\abs{\beta}H}-1)^2$ as opposed to $e^{2|\beta|}$. Despite the implicit dependence on $e^{|\beta| H}$, as $\sigma_h^2(s_h,a_h)\le (e^{\abs{\beta}H}-1)^2$ and $\lambda$ is adjusted correspondingly, we have $(\Sigma_h^*)^{-1}\preccurlyeq (e^{\abs{\beta}H}-1)^2 \Lambda_h^{-1}$, ensuring that $\norm{\phi(s,a)}_{(\Sigma_h^*)^{-1}}\le(e^{\abs{\beta}H}-1)\Lnorm{\phi(s,a)}$. In other words, Algorithm \ref{alg:alg2} is never worse than Algorithm \ref{alg:alg1} with appropriate parameters under Assumption \ref{assump:coverage}.

{\noindent \bf Comparison with Existing Results.} Finally, we discuss our results in the context of existing works on both risk-sensitive RL and offline RL. We begin by discussing our dependence on $\frac{e^{|\beta|H} - 1}{|\beta|}$, which we dub the risk-sensitive factor. All of our bounds have a dependence on $\tilde{O}(\frac{e^{|\beta|H} - 1}{|\beta|})$. According to \citet{fei2021exponential,fei2021risk,liang2022bridging}, in the online setting, the lower bound also depends on such a factor. We can only conjecture such a dependence might exist inspired by the lower bound for the online setting. However, it remains an open question how to derive a lower bound for risk-sensitive offline RL under linear MDPs. 
Compared with \citet{fei2021risk}, the only paper studying linear function approximation in risk-sensitive RL though for online setting, our bound can remove a potential factor of $e^{|\beta|H^2}$ by the shifting and scaling technique.  
Moreover, under a mild coverage assumption, our bound's dependence on $d$ can be improved to $\tilde{O}(\sqrt{d})$, improving over the result in \citet{fei2021risk} by a factor of $\sqrt{d}$. When compared to risk-neutral offline RL in linear MDPs, our dependence on $d$ matches the performance guarantee in~\citet{xiong2022nearly}, the tightest bounds for risk-neutral RL in linear MDPs that we are aware of, while also matching the lower bound for risk-neutral RL in the same paper. 

\section{Theoretical Analysis}
\label{sec:theoretical-analysis}
In this section, we outline the analysis of our theorems. Formal proofs are deferred to the appendix. Specifically, Appendix \ref{sec:proofs-for-thm1} presents the proof of Theorem \ref{thm1}, Appendix \ref{sec:proofs-for-thm2} provides the proof of Theorem \ref{thm2}, and Appendix \ref{sec:proofs-for-thm3} presents the proof of Theorem \ref{thm3}.

\subsection{Proof Sketch of Theorem \ref{thm1}} 
By Lemma \ref{lemma:pessimistic-bound-of-suboptimality}, the bound of the suboptimality can be obtained by bounding the model evaluation error, which is controlled with the estimation error of Bellman operator as shown in Lemma \ref{lemma3}. Taking into account the new structure exponential Bellman equation in \eqref{eq:ebellman}, in the risk-sensitive setting, the Bellman operators $\BB_h$ and its estimate $\wh\BB_h$ are defined as
\begin{align}\label{Bellman-operator}
    \BB_h f&= \begin{cases}
        e^{\beta (r_h-1)}(\PP_h f+e^{\beta h}),~ \beta >0\\
        e^{\beta r_h}(e^{-\beta H}-\PP_h f),\quad \beta <0
    \end{cases},\\
    \wh\BB_h f&= \begin{cases}
        e^{\beta (\wh r_h-1)}(\wh\PP_h f+e^{\beta h}),~ \beta >0\\
        e^{\beta \wh r_h}(e^{-\beta H}-\wh \PP_h f),\quad \beta <0
    \end{cases},
\end{align}
where $\wh \PP_h$ is an estimate of the transition operator. With the appropriate choice of $\lambda$, the dominating term in the estimation error $\wh\BB_h (\SSS_{h+1}\wV_{h+1})(s,a)-\BB_h (\SSS_{h+1}\wV_{h+1})(s,a)$ is $\big\|\sum_\tau \phi_h^\tau \epsilon_h^\tau(\SSS_{h+1}\wV_{h+1})\big\|_{\invL}\Lnorm{\phi}$ where the term $\epsilon_h^\tau(\SSS_{h+1}\wV_{h+1})$ is the regression noise defined by $\SSS_{h+1}\wV_{h+1}(s_{h+1}^\tau) - \PP_h( \SSS_{h+1}\wV_{h+1})(s_h^\tau,a_h^\tau)$, which can in turn be controlled by a uniform-concentration bound over the possible values of $\SSS_{h+1}\wV_{h+1}$. We remark that obtaining a uniform concentration of the term requires careful analysis of the covering number of the function class $\SSS_{h+1}\wV_{h+1}$ for risk-sensitive linear MDPs, which has not been examined prior to our work. Fortunately, the ``shifting and scaling'' operator $\SSS$ helps ensure that $\SSS_{h}\wV_{h}$ is scaled similarly at each $h$, streamlining our analysis. By a combination of ``shifting and scaling'' and uniform concentration analysis, we are able to relate the model evaluation error in risk-sensitive MDPs to the risk-neutral uncertainty quantifier $\gamma \Lnorm{\phi(\cdot, \cdot)}$, which completes our proof.

\subsection{Proof Sketch of Theorem \ref{thm2}}
We sharpen Algorithm \ref{alg:alg1}'s dependence on $d$ in Theorem \ref{thm2} under Assumption \ref{assump:coverage} by better adjusting $\gamma$, yet without changing the algorithm's overall structure. An additional $\sqrt{d}$ amplification of the error upper bound is introduced in the proof of Theorem~\ref{thm1} due to the uniform concentration analysis. 
We further show that when the data has sufficient coverage, the performance of offline risk-sensitive RL algorithms can benefit from reference-advantage decomposition by avoiding this amplification. The main idea is to set a fixed reference function $V_{h+1}^{\rm ref}$ and decompose the error of the transition $\PP_h(\SSS_{h+1}\wV_{h+1}) -\wh\PP_h (\SSS_{h+1}\wV_{h+1})$ as $\PP_h(\SSS_{h+1}V^{\rm ref}_{h+1}) -\wh\PP_h (\SSS_{h+1}V^{\rm ref}_{h+1})$ and $\PP_h(\SSS_{h+1}\wV_{h+1}-\SSS_{h+1}V^{\rm ref}_{h+1}) -\wh\PP_h (\SSS_{h+1}\wV_{h+1}-\SSS_{h+1}V^{\rm ref}_{h+1})$.

As $V_{h+1}^{\rm ref}$ is fixed, we avoid the $\sqrt{d}$ amplification in the first term by eschewing uniform concentration. For the second term, if $\big\|\SSS_{h+1}\wV_{h+1}-\SSS_{h+1}V^{\rm ref}_{h+1}\big\|_\infty\le R_{h+1}$ for some constant $R_{h+1}$, the uniform concentration as in Theorem \ref{thm1} leads to an upper bound $\tilde O(dR_{h+1})\Lnorm{\phi}$ which will be non-dominating as long as $R_{h+1}$ is sufficiently small. Equipped with data coverage assumption, Assumption \ref{assump:coverage}, the optimal value function $V^*_{h}$ may serve as the reference function and $\big\|\SSS_hV^*_h-\SSS_h\wV_h\big\|_\infty$ is small enough for sufficiently large $K$, as the dataset now has sufficient coverage of the optimal risk-sensitive policy. Then we obtain a tighter bound on the error $\wh\BB_h (\SSS_{h+1}\wV_{h+1})(s,a)-\BB_h (\SSS_{h+1}\wV_{h+1})(s,a)$. Correspondingly, setting the parameter $\gamma$ in Algorithm \ref{alg:alg1} as $\gamma = \tilde O(\sqrt{d}(e^{\abs{\beta}H}-1))$ provides a tighter bound on the suboptimality. 

\subsection{Proof Sketch of Theorem \ref{thm3}} 

We can further sharpen the dependency on the risk-sensitivity factor $\frac{e^{|\beta|H} - 1}{|\beta|}$ via variance information. Recall from the proof sketch of Theorem~\ref{thm1} that the key quantity is $\Snorm{\sum_\tau \frac{\phi_h^\tau}{\wh\sigma_h(s_h^\tau,a_h^\tau)} \tilde\epsilon_h^\tau(\SSS_{h+1}V^*_{h+1})}$ with the weighted noise $\tilde\epsilon_h^\tau(\SSS_{h+1}V^*_{h+1}) = \epsilon_h^\tau(\SSS_{h+1}V^*_{h+1})/\wh\sigma_h(s_h^\tau,a_h^\tau)$. Intuitively, since the variance of the noise term is not necessarily the same for all $(s, a)$, weighing each observation accordingly improves the bound via the Gauss-Markov theorem~\citep{amemiya1985advanced}. When data has sufficient coverage and the number of samples is sufficiently large, the variance can be accurately estimated, thereby improving the bounds. 

We sketch our technique below. Let $\sigma^2$ and $R$ denote the conditional variance and magnitude of $\tilde \epsilon_h^\tau$. We use a Bernstein-type concentration inequality and improve the term's bound to $\tilde O(\sqrt{d}\sigma+R)$, whereas the standard Hoeffding-type used to prove Theorem \ref{thm1} yields $\tilde O(\sqrt{d}R)$. Thus, as long as the estimator $\wh\sigma_h^2$, defined in~\eqref{eq:variance-estimator}, is consistent, with an appropriate choice of $\underline\sigma^2$, we have $R=O(\sqrt{d})$ and $\sigma = O(1)$ leading to the upper bound in Theorem \ref{thm3}.

To show the consistency of $\wh\sigma_h^2$, we first bound the estimation error between $\wh\sigma_h^2$ and $\max\{\underline\sigma^2,{\rm Var }_h(\SSS_{h+1}\wV_{h+1})(s,a)\}$ by similar technique as in Theorem \ref{alg:alg1}. Then we can convert $\rm{Var}_h(\SSS_{h+1}\wV_{h+1})(s,a)$ to ${\rm Var}_h(\SSS_{h+1}V^*_{h+1})(s,a)$ under the coverage assumption, since $\SSS_{h+1}\wV_{h+1}$ and $\SSS_{h+1}V^*_{h+1}$ are close enough for large $K$. 

\section{Conclusion}
We study risk-sensitive offline reinforcement learning under the entropic risk measure, with a focus on the linear MDP. We begin by presenting a risk-sensitive pessimistic value iteration algorithm, offering a tight analysis by leveraging the structure of the risk-sensitive performance measure. To further improve the obtained bounds, we propose another pessimistic algorithm that utilizes variance information and reference-advantage decomposition, improving both the dependence on the space dimension $d$ and the risk-sensitivity factor. To the best of our knowledge, we obtain the first provably efficient risk-sensitive offline RL algorithms.

\section*{Acknowledgements}

The authors would like to thank the reviewers for their valuable feedback.

\section*{Impact Statement}
While motivated by practical applications, our work is theoretical in nature. Particularly, our results concern the theoretical guarantees for provably efficient offline RL algorithms and constitute generic algorithmic and theoretical contributions to reinforcement learning. In our research, we do not investigate specific application scenarios. The societal impact of our algorithms depends on how practitioners employ them and what scenarios they apply our algorithms to. Overall, this paper presents work whose goal is to advance the field of machine learning, especially from a theoretical perspective. There are many potential societal consequences of our work, none of which we feel must be specifically highlighted here.

\bibliography{reference}
\bibliographystyle{icml2024}


\newpage
\appendix
\onecolumn
{\centering
{\LARGE Appendix}
\par }


\addtocontents{toc}{\protect\setcounter{tocdepth}{2}}

\section{Proof of Lemma~\ref{lemma:pessimistic-bound-of-suboptimality}}
\label{sec:proof-of-lemma-pessimistic-bound-of-suboptimality}
\begin{proof}
      As $\wh\pi$ is the greedy policy, we have $\widehat V_h(s) =\frac{1}{\beta}\log (\inner{e^{\beta\wQ_h(s,\cdot)},\wh\pi_h(\cdot|s)}_\cA)$. Then, for any policy $\pi^\prime$, using $e^{\beta V_h^{\pi'}(s)} =  \inner{e^{\beta Q^{\pi'}_h(s,\cdot)},\pi'_h(\cdot|s)}_\cA=\inner{e^{\beta r_h(s,\cdot)}\PP_he^{\beta V^{\pi^\prime}_{h+1}}(s,\cdot),\pi'_h(\cdot|s)}_\cA$ from \eqref{eq:ebellman}, we have
\begin{align}
\begin{aligned}\label{l3.1:eqn1}
        e^{\beta V_h^{\pi^\prime}(s)}- e^{\beta\wV_h(s)} =&\inner{e^{\beta\wQ_h(s,\cdot)},\pi_h^\prime(\cdot|s)-\wh\pi_h(\cdot|s)}_\cA +  \inner{\iota_{\mathrm{exp},h}(s,\cdot),\pi_h^\prime(\cdot|s)}_\cA \\
    &+ \inner{e^{\beta r_h(s,\cdot)}(\PP_he^{\beta V^{\pi^\prime}_{h+1}}(s,\cdot)-\PP_h e^{\beta\wV_{h+1}}(s,\cdot)),\pi_h^\prime(\cdot|s)}_\cA,
\end{aligned}
\end{align}
with $\iota_{\mathrm{exp},h}(s,a) = e^{\beta r_h(s,a)}\PP_h e^{\beta\wV_{h+1}}(s_h,a_h)-e^{\beta\wQ_h(s,a)}$. Then, we consider the case $\beta>0$ and $\beta<0$ separately. 

\textbf{Case 1: $\beta>0$.} 
 As $\wh\pi_h(\cdot|s)= \argmax_\pi \inner{\wh Q_h(s,\cdot),\pi(\cdot|s)}_\cA$, we have $\inner{e^{\beta\wQ_h(s,\cdot)},\pi_h^\prime(\cdot|s)-\wh\pi_h(\cdot|s)}_\cA\le0$. Given this property, if we set $\pi^\prime=\pi$ in \eqref{l3.1:eqn1}, we have
\begin{align}
\begin{aligned}\label{l3.1:eqn2}
    &e^{\beta V_h^{*}(s_h)}-e^{\beta\wV_h(s_h)}  \\
    &~~ \le \EE_{\pi^*}\left[\iota_{\mathrm{exp},h}(s_h,a_h)|s_h\right] + \EE_{\pi^*}\left[e^{\beta r_h(s_h,a_h)}(\PP_h e^{\beta V^{*}_{h+1}}(s_h,a_h)-\PP_he^{\beta \wV_{h+1}}(s_h,a_h))\Big|s_h\right].
\end{aligned}
\end{align}
If we set $\pi^\prime = \wh\pi$ in \eqref{l3.1:eqn1}, we have
\begin{align}
\begin{aligned}\label{l3.1:eqn3}
    &e^{\beta V_h^{\wh\pi}(s_h)}-e^{\beta\wV_h(s_h)}  \\
    &~~=\EE_{\wh\pi}\left[\iota_{\mathrm{exp},h}(s_h,a_h)|s_h\right] + \EE_{\wh\pi}\left[e^{\beta r_h(s_h,a_h)}(\PP_h e^{\beta V^{\wh\pi}_{h+1}}(s_h,a_h)-\PP_he^{\beta \wV_{h+1}}(s_h,a_h))\Big|s_h\right].
\end{aligned}
\end{align}
Equipped with \eqref{l3.1:eqn2} and \eqref{l3.1:eqn3}, we can show the following \eqref{l3.1:eqn4} by induction from $h=H$ to $h=1$,
\begin{align}\label{l3.1:eqn4}
    \begin{aligned}
     &e^{\beta V_h^{*}(s_h)}-e^{\beta\wV_h(s_h)}\le \sum_{h'=h}^H e^{\beta(h'-h)}\EE_{\pi^*}\left[\iota_{\mathrm{exp},h'}(s_{h'},a_{h'})|s_h\right],\\
     &e^{\beta V_h^{\wh\pi}(s_h)}-e^{\beta\wV_h(s_h)}\ge 0.
\end{aligned}
\end{align}
To see this, we start with the base case $h=H$. As $\wV_{H+1}=V^{\wh\pi}_{H+1} = V^*_{H+1}=0$, \eqref{l3.1:eqn4} is implied by \eqref{l3.1:eqn2} and \eqref{l3.1:eqn3} directly with the assumption $\iota_{\mathrm{exp},H}\ge0$. Supposing that \eqref{l3.1:eqn2} holds for $h'\ge h+1$, we aim to show that it also holds for $h$. In fact, by \eqref{l3.1:eqn3}, $\iota_{\mathrm{exp},h}\ge0$ and the induction assumption, we have
\begin{align}\label{l3.1:eqn5}
    e^{\beta V_h^{\wh\pi}(s_h)} - e^{\beta \wV_h(s_h)} \ge  \EE_{\wh\pi}\left[\iota_{\mathrm{exp},h}(s_h,a_h)|s_h\right] \ge 0.
\end{align}
In addition, as $e^{\beta V_{h+1}^{*}(s_{h+1})}-e^{\beta\wV_{h+1}(s_{h+1})}\ge 0 $, \eqref{l3.1:eqn2} leads to
\begin{align}
    \begin{aligned}\label{l3.1:eqn6}
    &e^{\beta V_h^{*}(s_h)}-e^{\beta\wV_h(s_h)}   \\
    \le& \EE_{\pi^*}\left[\iota_{\mathrm{exp},h}(s_h,a_h)|s_h\right] + e^\beta\EE_{\pi^*}\left[\PP_h e^{\beta V^{*}_{h+1}}(s_h,a_h)-\PP_he^{\beta \wV_{h+1}}(s_h,a_h)\Big|s_h\right]\\
    \le& \EE_{\pi^*}\left[\iota_{\mathrm{exp},h}(s_h,a_h)|s_h\right] + e^\beta\EE_{\pi^*}\left[\sum_{h'=h+1}^H e^{\beta(h'-h-1)}\EE_{\pi^*}\left[\iota_{\mathrm{exp},h'}(s_{h'},a_{h'})|s_{h+1}\right]\Big|s_h\right]\\
    =& \sum_{h'=h}^H e^{\beta(h'-h)}\EE_{\pi^*}\left[\iota_{\mathrm{exp},h'}(s_{h'},a_{h'})|s_h\right],
\end{aligned}
\end{align}
where the first inequality follows from $r_h\in[0,1]$ and the second inequality follows from the induction assumption that \eqref{l3.1:eqn4} holds for $h+1$. By induction, \eqref{l3.1:eqn5} and \eqref{l3.1:eqn6} imply that \eqref{l3.1:eqn4} holds for any $h\in[H]$. Moreover, setting $h=1$ in \eqref{l3.1:eqn4}, we have
\begin{align*}
    e^{\beta V_1^{*}(s_1)}-e^{\beta V_1^{\wh\pi}(s_1)}
        \le e^{\beta V_1^{*}(s_1)}-e^{\beta \wV_1(s_1)}
        \le&\sum_{h=1}^He^{\beta (h-1)}\EE_{\pi^*}\left[\iota_{\mathrm{exp},h}(s_h,a_h)|s_1\right].
\end{align*}
Together with Lemma \ref{lemma1}, we have
\begin{align*}
    \mathrm{SubOpt}(\wh\pi) = V_1^{*}(s_1)-V_1^{\wh\pi}(s_1)\le\frac{1}{\beta}\left(e^{\beta V_1^{*}(s_1)}-e^{\beta V_1^{\wh\pi}(s_1)}\right) \leq \sum_{h=1}^H\frac{e^{\beta (h-1)}}{\beta}\EE_{\pi^*}\left[\iota_{\mathrm{exp},h}(s_h,a_h)|s_1\right].
\end{align*}
Therefore, we conclude the proof of the case $\beta>0$.

\textbf{Case 2: $\beta<0$.} Similar to the positive $\beta$ case, by setting $\pi^\prime=\pi$ and $\pi'=\wh\pi$ in \eqref{l3.1:eqn1} and the assumption $\iota_{\mathrm{exp},h}\le0$, we have 
\begin{align*}
\begin{aligned}
    &e^{\beta V_h^{*}(s_h)}-e^{\beta\wV_h(s_h)}  \\
    &\ge \EE_{\pi^*}\left[\iota_{\mathrm{exp},h}(s_h,a_h)|s_h\right] + \EE_{\pi^*}\left[e^{\beta r_h(s_h,a_h)}(\PP_h e^{\beta V^{*}_{h+1}}(s_h,a_h)-\PP_he^{\beta \wV_{h+1}}(s_h,a_h))\Big|s_h\right]
\end{aligned}
\end{align*}
and
\begin{align*}
\begin{aligned}
    &e^{\beta V_h^{\wh\pi}(s_h)}-e^{\beta\wV_h(s_h)}  \le\EE_{\wh\pi}\left[e^{\beta r_h(s_h,a_h)}(\PP_h e^{\beta V^{\wh\pi}_{h+1}}(s_h,a_h)-\PP_he^{\beta \wV_{h+1}}(s_h,a_h))\Big|s_h\right].
\end{aligned}
\end{align*}
Again, by induction from $h=H$ to $h=1$, we can obtain a result similar to \eqref{l3.1:eqn4},
\begin{align}\label{l3.1:eqn9}
    \begin{aligned}
     &e^{\beta V_h^{*}(s_h)}-e^{\beta\wV_h(s_h)}\ge \sum_{h'=h}^H \EE_{\pi^*}\left[\iota_{\mathrm{exp},h'}(s_{h'},a_{h'})|s_h\right],\\
     &e^{\beta V_h^{\wh\pi}(s_h)}-e^{\beta\wV_h(s_h)}\le 0,
\end{aligned}
\end{align}
and \eqref{l3.1:eqn9} leads to
\begin{align*}
    e^{\beta V_1^{*}(s_1)}-e^{\beta V_1^{\wh\pi}(s_1)}
        \ge e^{\beta V_1^{*}(s_1)}-e^{\beta \wV_1(s_1)}
        \ge&\sum_{h=1}^H\EE_{\pi^*}\left[\iota_{\mathrm{exp},h}(s_h,a_h)|s_1\right].
\end{align*}
Together with Lemma \ref{lemma1}, as $\beta<0$, we have 
\begin{align*}
       \mathrm{SubOpt}(\wh\pi)\le\frac{e^{-\beta H}}{\beta}\left(e^{\beta V_1^{*}(s_1)}-e^{\beta V_1^{\wh\pi}(s_1)}\right) \le \frac{e^{-\beta H}}{\beta}\sum_{h=1}^H\EE_{\pi^*}\left[\iota_{\mathrm{exp},h}(s_h,a_h)|s_1\right].
\end{align*}
The proof of the case $\beta<0$ is concluded.
\end{proof} 

\section{Proof of Theorem \ref{thm1}}
\label{sec:proofs-for-thm1}

In this section, we provide the detailed proofs for results attained by Algorithm~\ref{alg:alg1} with $\gamma =\tilde O( d(e^{\abs{\beta}H}-1))$. Furthermore, this section will serve as a warm-up for the subsequent section in which we sharpen the upper bound.

 \begin{proof}[Proof of Theorem~\ref{thm1}]
We begin with the following result, which relates suboptimality to the transformed space induced by $\SSS_h$, the shifting and scaling operator. 
\begin{lemma}\label{lemma1}
    We have 
    \begin{align*}
        V_1^*(s_1) - V_1^{\wh \pi}(s_1)\le \frac{1}{\beta}(e^{\beta V_1^*(s_1)}-e^{\beta V_1^{\wh \pi}(s_1)}) = \frac{1}{\abs\beta}(\SSS_1 V_1^*(s_1)-\SSS_1 V_1^{\wh\pi}(s_1))
    \end{align*}
    for all $\beta>0$ and
    \begin{align*}
        V_1^*(s_1) - V_1^{\wh \pi}(s_1) \le \frac{e^{-\beta H}}{\beta}(e^{\beta V_1^*(s_1)}-e^{\beta V_1^{\wh \pi}(s_1)})=\frac{1}{\abs\beta}(\SSS_1 V_1^*(s_1)-\SSS_1 V_1^{\wh\pi}(s_1))
    \end{align*}
    for all $\beta<0$.
\end{lemma}
The lemma then permits us to work almost entirely in the space induced by $\SSS_h$ for the rest of the proof, which is equipped with the property that $\SSS_h V_h$ is roughly of the same scale at all $h \in [H]$, thereby avoiding the need to adjust uniform concentration bounds at each step.
We then define the model evaluation error after the transformation $\SSS_h$ as 
  \begin{align*}
        \iota_h(s_h,a_h) =\begin{cases}
             e^{\beta(h-1)}\iota_{exp,h}(s_h,a_h), \quad \beta >0\\
             -e^{-\beta H}\iota_{exp,h}(s_h,a_h),\quad\beta <0
        \end{cases}
    \end{align*}
The following lemma captures the relationship between suboptimality and the error. It can be viewed as a new version of Lemma \ref{lemma:pessimistic-bound-of-suboptimality} after introducing $\iota_h$ and $\SSS_h$, in the space induced by $\SSS_h$.
\begin{lemma}\label{lemma2}
If $\iota_h(s_h,a_h)\ge 0$ for all $s_h, a_h,$ and $h$, then
\begin{align}\label{l2:eqn0}
    &\SSS_h V_h^*(s_h) - \SSS_h \wh V_h(s_h) \nonumber\\
    &\qquad \le \EE_{\pi^*}[\iota_h(s_h,a_h)|s_h] + \EE_{\pi^*}[\SSS_{h+1} V_{h+1}^*(s_{h+1}) - \SSS_{h+1} \wh V_{h+1}(s_{h+1})|s_h],
\end{align}
    and
    \begin{align}\label{l2:eqn0.1}
        \SSS_h V_h^{\wh\pi}(s_h) - \SSS_h \wh V_h(s_h) \ge0
    \end{align}
    Consequently, it holds that
        \begin{align*}
        \SSS_1 V_1^{*}(s_1) - \SSS_1 V^{\wh\pi}_h(s_1) \le \sum_{h=1}^H\EE_{\pi^*}[\iota_h(s_h,a_h)|s_1] .
    \end{align*}
\end{lemma}

The result then shows that if the estimate is pointwise pessimistic, in the sense that $\iota_h(s_h, a_h) \geq 0$ for all $s_h, a_h, h$, then the suboptimality upper bound defined in Lemma~\ref{lemma2} can be in turn translated to the expected model evaluation error, where expectation is taken over the state action visitation measure induced by the optimal policy. Naturally, the next step is to ensure that the condition $\iota_{h}(s_h, a_h) \geq 0$ holds and controlling the term from above via $\Gamma_h(s, a)$, the uncertainty bonus defined in Algorithm~\ref{alg:alg1}.

We first provide a more formal definition of the operators defined in \eqref{Bellman-operator} in the proof sketch. Particularly, let
\begin{align}\label{eqn:operator-B}
    \BB_h f= \begin{cases}
        e^{\beta (r_h-1)}(\PP_h f+e^{\beta h}),~ \beta >0\\
        e^{\beta r_h}(e^{-\beta H}-\PP_h f),\quad \beta <0
    \end{cases}
    ~~\text{and}\quad\wh\BB_h f= \begin{cases}
        e^{\beta (\wh r_h-1)}(\wh\PP_h f+e^{\beta h}),~ \beta >0\\
        e^{\beta \wh r_h}(e^{-\beta H}-\wh \PP_h f),\quad \beta <0
    \end{cases},
\end{align}
where $\wh\PP_h f(s,a) = \phi(s,a)^\top \wh w(f)$ with
\begin{align}\label{eqn:phat}
    \wh w(f) = \Sigma_h^{-1}\left(\sum_{\tau}\frac{\phi_h^\tau \cdot f(s_{h+1}^\tau)}{\wh\sigma_h^2(s_h^\tau,a_h^\tau)}\right)~\text{ and }~
    \Sigma_h = \sum_{\tau}\frac{\phi_h^\tau (\phi_h^\tau)^\top}{\wh\sigma_h^2(s_h^\tau,a_h^\tau)}+\lambda\cdot I.
\end{align}

We use $\phi_h^\tau$ to denote $\phi(s_h^\tau,a_h^\tau)$ for notation simplicity. With the definition in mind, we then show that conditioned on a so-called ``good event'', the term $\iota_h$ can be controlled simultaneously from above and below.

\begin{lemma}\label{lemma3}
On the event
\begin{align*}
    \cE_h=\left\{\abs{\BB_h  (\SSS_{h+1}\wV_{h+1})(s,a) - \wh\BB_h  (\SSS_{h+1}\wV_{h+1})(s,a) }\le \Gamma_h(s,a)\right\},
\end{align*}
we have $0\le\iota_h(s,a)\le 2 \Gamma_h(s,a)$.
\end{lemma}

Combining Lemma \ref{lemma2} and \ref{lemma3}, we have 
\begin{align}\label{thm1:eqn1}
    V_1^*(s_1) - \wV_1(s_1)\le \frac{1}{\abs\beta}(\SSS_h V_1^*(s_1) - \SSS_h \wV_1(s_1)) \le \frac{2}{\abs{\beta}}\sum_{h=1}^H\EE_{\pi^*}[\Gamma_h(s_h,a_h)|s_1]
\end{align}
for some bonus function $\Gamma_h(s,a)$ if 
\begin{align}\label{thm1:eqn2}
    \cE_h=\left\{\abs{\BB_h  (\SSS_{h+1}\wV_{h+1})(s,a) - \wh\BB_h  (\SSS_{h+1}\wV_{h+1})(s,a) }\le \Gamma_h(s,a)\right\}
\end{align}
holds for any $h$. In Algorithm \ref{alg:alg1}, the bonus function is set to be $\Gamma_h(s,a) =\gamma \Lnorm{\phi(s,a)}$. Then, to prove the theorem, it is sufficient to show that $\PP(\cap_{h=1}^H\cE_h)\ge 1-\delta$ with $\gamma =\tilde O(dR_\beta)$ where we use $R_\beta$ to denote $e^{\abs{\beta}H}-1$. \eqref{thm1:eqn2} suggests us to focus on the error between $\BB_h(\SSS_{h+1}\wV_{h+1})$ and $\wh\BB_h(\SSS_{h+1}\wV_{h+1})$. We first provide an upper bound on the error that is easier to control
\begin{lemma}\label{lemma5}
The following bound holds for all $\beta, s, a, $ and $h$.
 \begin{align}
\begin{aligned}\label{thm1:eqn3}
        &\abs{\BB_h (\SSS_{h+1}\wV_{h+1})(s,a) -\wh\BB_h (\SSS_{h+1}\wV_{h+1})(s,a)}\\
        &\le e^{\abs{\beta} H}\abs{\beta} \underbrace{\abs{r_h(s,a)-\wh r_h(s,a)}}_{\textrm{(i)}} +\underbrace{\abs{\PP_h (\SSS_{h+1}\wV_{h+1})(s,a) -\wh\PP_h (\SSS_{h+1}\wV_{h+1})(s,a)}}_{\textrm{(ii)}}.
\end{aligned}    
\end{align}
\end{lemma}
Here, $\textrm{(i)}$ is the error we incur when estimating the reward function $r_h$ from data and $\textrm{(ii)}$ from estimating the transition operator, $\PP_h$, in the space induced by $\SSS_{h + 1}$. Both can be controlled simultaneously by the following lemma:
\begin{lemma}\label{lemma4}
For a function $\norm{g_{h+1}}_\infty\le R$ with $\PP_h g_{h+1}(s,a)=\phi(s,a)^\top w_h$, we have 
\begin{align*}
    \abs{\PP_h g_{h+1}(s,a) -\wh\PP_h g_{h+1}(s,a)} \le&  R\sqrt{d\lambda} \Snorm{\phi(s,a)}\\&+ \Snorm{\sum_\tau \frac{\phi_h^\tau }{\wh\sigma_h(s_h^\tau,a_h^\tau)}\cdot \tilde\epsilon_h^\tau(g_{h+1})} \Snorm{\phi(s,a)},
\end{align*}
where $\tilde\epsilon_h^\tau(g) = \frac{g(s_{h+1}^\tau) - \PP_h g(s_h^\tau,a_h^\tau)}{\wh\sigma_h(s_h^\tau,a_h^\tau)}$ and $\wh\PP_h$ is defined in \eqref{eqn:phat}. In addition, we have
\begin{align*}
    \abs{r_h(s,a) - \wh r_h(s,a)}\le \sqrt{d\lambda}\Snorm{\phi(s,a)}
\end{align*}
for $\wh r_h(s,a)$ given in Algorithms \ref{alg:alg1} and \ref{alg:alg2}.
\end{lemma}
We note that the lemma is stated (and proven) in a more general fashion such that it can be applied to both Algorithm~\ref{alg:alg1} and Algorithm~\ref{alg:alg2}. To specialize the result to Theorem~\ref{thm1}, all we need is to set $\wh\sigma^2_h(s,a)=1$ for all $s, a,$ and $h$, thereby obtaining 
\begin{align}\label{thm1:eqn4}
    (i)=\abs{r_h(s,a)-\wh r_h(s,a)}\le \sqrt{d\lambda}\Lnorm{\phi(s,a)}.
\end{align}
In addition, as $\norm{\SSS_{h+1}\wV_{h+1}}_\infty\le R_\beta:= e^{\abs{\beta}H}-1$, Lemma \ref{lemma4} also leads to 
\begin{align}\label{thm1:eqn5}
    \textrm{(ii)} \le  R_\beta\sqrt{d\lambda} \Lnorm{\phi(s,a)}+ \Lnorm{\sum_\tau \phi_h^\tau\cdot \epsilon_h^\tau(\SSS_{h+1}\wV_{h+1})} \Lnorm{\phi(s,a)},
\end{align}
where the first term is caused by the bias of ridge regression and $\epsilon_h^\tau(g) = g(s_{h+1}^\tau) - \PP_h g(s_h^\tau,a_h^\tau)$. Plugging \eqref{thm1:eqn4} and \eqref{thm1:eqn5} into  \eqref{thm1:eqn3}, we have
\begin{align}
\begin{aligned}\label{thm1:eqn6}
    &\abs{\BB_h (\SSS_{h+1}\wV_{h+1}) -\wh\BB_h (\SSS_{h+1}\wV_{h+1})}\\
    &\le  (R_\beta+e^{\abs{\beta}H}\abs{\beta}) \sqrt{d\lambda}\Lnorm{\phi(s,a)} + \Lnorm{\sum_\tau \phi_h^\tau \epsilon_h^\tau(\SSS_{h+1}\wV_{h+1})}\Lnorm{\phi(s,a)}\\
    &\le 2R_\beta \sqrt{d}\Lnorm{\phi(s,a)} + \Lnorm{\sum_\tau \phi_h^\tau \epsilon_h^\tau(\SSS_{h+1}\wV_{h+1})}\Lnorm{\phi(s,a)}.
\end{aligned}
\end{align}
The last step in \eqref{thm1:eqn6} follows the fact $\sqrt{\lambda} = e^{-\abs{\beta}}$ and the inequality
\begin{align}\label{thm1:eqn7}
e^{\abs{\beta}H}\abs{\beta}\le e^{\abs{\beta}H}(e^{\abs{\beta}}-1)\le e^{\abs{\beta}(H+1)}-e^{\abs{\beta}} = \frac{R_\beta}{\sqrt{\lambda}} .
\end{align}
With \eqref{thm1:eqn6}, it is sufficient to bound $ \Lnorm{\sum_\tau \phi_h^\tau \epsilon_h^\tau(\SSS_{h+1}\wV_{h+1})} $ as the first term is already $\tilde O(dR_\beta )\Lnorm{\phi(s,a)}$. As we do not have any coverage assumption on the offline dataset for Theorem~\ref{thm1}, we do so via uniform concentration.

\textbf{Uniform Concentration.} In the backward iteration, $\wh V_{h+1}$ depends on the data $(s_{h+1}^\tau, a_{h+1}^\tau)$ and thus a uniform concentration result for $ \Lnorm{\sum_\tau \phi_h^\tau \epsilon_h^\tau(\SSS_{h+1}\wV_{h+1})} $ is needed. For brevity, in this section, we focus on the case $\beta>0$ first and consider the following function class
\begin{align*}
    &\cU_h(L_\theta,L_w,L_\gamma,\lambda) \\
    &\qquad = \left\{U_h(s;\theta,w,\gamma,\Sigma):\cS\to [0,R_\beta] \textrm{ s.t. } \norm{\theta}\le L_\theta, \norm{w}\le L_w, \gamma\in[0, L_\gamma], \Sigma \gtrsim \lambda\cdot I\right\},
\end{align*}
where 
\begin{align*}
        &U_h(s;\theta,w,\gamma,\Sigma)\\
        &\qquad =\max_{a}\left\{e^{\beta[\{\phi(s,a)^\top\theta\}_{[0,1]}-1]}(\phi(s,a)^\top w +e^{\beta h})-\gamma \sqrt{\phi(s,a)^\top \Sigma^{-1}\phi(s,a)}\right\}_{[e^{\beta(h-1)},e^{\beta H}]}.
\end{align*}
Given the construction of $\wV_h$ in Lines 6-9 of Algorithm \ref{alg:alg1}, it is straightforward to see that
$e^{\beta(\wV_{h}+h-1)}\in \cU_h(L_\theta,L_w,L_\gamma,\lambda)$. Controlling the size of $\cU_h$ relies on controlling the norms of the estimates, which we provide in the following lemma.
 \begin{lemma}[Upper bound of estimated coefficients]\label{lemma:coef-upper-bound} $\wh w_h$ and $\theta_h$ from Algorithm \ref{alg:alg1} satisfy
 \begin{align*}
     \norm{\wh w_h}\le R_\beta\sqrt{\frac{Kd}{\lambda}},~~~\norm{\wh \theta_h}\le \sqrt{\frac{Kd}{\lambda}},
 \end{align*}
 where $R_\beta = e^{\abs{\beta}H}-1$. In addition, $\wh w_h$ and $\theta_h$ from Algorithm \ref{alg:alg2} satisfy
 \begin{align*}
     \norm{\wh w_h}\le R_\beta^2\sqrt{\frac{Kd}{\lambda}},~~~\norm{\wh \theta_h}\le R_\beta\sqrt{\frac{Kd}{\lambda}}.
 \end{align*}
\end{lemma}
Using the upper bounds of $\norm{\wh w_h}$ and $\norm{\wh\theta_h}$ in Lemma \ref{lemma:coef-upper-bound} and the fact that $\gamma$ is set to be $\gamma =c\cdot dR_\beta \sqrt{\zeta}$ with $\zeta =\log{(3 dH K e^{\abs{\beta}}/\delta)}$ and some constant $c$, we have 
\begin{align}\label{thm1:eqn8}
    L_\theta = \sqrt{Kd/\lambda},~ L_w = R_\beta\sqrt{Kd/\lambda},~ L_\gamma = c\cdot d R_\beta\sqrt{\zeta},~\lambda = e^{-2\beta}.
\end{align}
As $\SSS_{h}\wV_h = e^{\beta(\wV_{h}+h-1)}-e^{\beta(h-1)}$, we have 
\begin{align}\label{thm1:eqn8.5}
    \SSS_{h}\wV_h\in \tilde\cU_{h}:=\left\{U_h -e^{\beta(h-1)}:U_h\in \cU_h(L_\theta,L_w,L_\gamma,\lambda)\right\}.
\end{align}
Let $\cN_h(\varepsilon)$ be the $\varepsilon$-cover of $\tilde\cU_h$ with respect to $\norm{\cdot}_\infty$ and $\abs{\cN_h(\varepsilon)}$ is the $\epsilon$-covering number. Thanks to the shifting and scaling technique, the $\epsilon$-covering number of $\cU_h$ and $\tilde\cU_h$ should be the same, and we provide the result in the following lemma.

\begin{lemma}[Upper bound of the covering number]\label{lemma:covering-number}
 For any $\varepsilon>0$, let $\abs{\cN_h(\varepsilon)}$ and $\abs{\cN'_h(\varepsilon)}$ be the $\varepsilon$-covering number of the function space $\cU_h(L_\theta,L_w,L_\gamma,\lambda)$ and $\cU'_h(L_\theta,L_w,L_\gamma,\lambda)$ respectively, we have
\begin{align*}
     &\max\{\log(\abs{\cN_h(\varepsilon)}),\log(\abs{\cN'_h(\varepsilon)})\}\\ &\qquad\le d\log(1+8L_w /\varepsilon) + d\log(1+8\abs{\beta} L_\theta (L_w+e^{\abs{\beta} H})/\varepsilon) + d^2\log(1+8d^{1/2}L_\gamma^2/(\lambda \varepsilon^2)).
\end{align*}
\end{lemma}
Here $\cN'_h(\varepsilon)$ refers to the covering number of the function class when $\beta < 0$. While we include its bound here for completeness, the term is defined formally in the proof of Lemma~\ref{lemma:covering-number} in Appendix~\ref{proof:lemma-covering-number}.

Setting $\varepsilon = \sqrt{\lambda}d R_\beta/K$, we have the following upper bound on the covering number.
\begin{align}
    \log \abs{\cN_h(\varepsilon)} &\le d\log(1+8L_w /\varepsilon) + d\log(1+8\beta L_\theta (L_w+e^{\beta H})/\varepsilon) + d^2\log(1+8d^{1/2}L_\gamma^2/(\lambda \varepsilon^2))\nonumber\\
    &\le d\log(1+8 e^{\beta} R_\beta\sqrt{Kd}/\varepsilon )\nonumber\\
    &\quad +d\log(1+16 e^{3\beta} R_\beta Kd/\varepsilon )+d^2\log(1+8c^2e^{2\beta}R_\beta^2 d^{5/2} \zeta/\varepsilon^2)\nonumber\\
    &= d\log(1+8 e^{2\beta} K^{3/2}d^{-1/2} )+d\log(1+16 e^{4\beta} K^2)+d^2\log(1+8c^2e^{4\beta} d^{1/2} K^2\zeta)\nonumber\\
    &\le 3d^2\log(32c^2e^{4\beta}d^{1/2}K^2\zeta). \label{thm1:eqn9}
\end{align}
Here, the first step comes from Lemma \ref{lemma:covering-number}. The second uses the parameters in \eqref{thm1:eqn8} and $e^{\beta H}-e^{\beta h}\le R_\beta$ and $e^{\beta}R_\beta\ge e^{\beta H}\beta$ from \eqref{thm1:eqn7}. The third step uses our choice of $\varepsilon$. The last step holds as long as $c\ge 1$. 

Equipped with the upper bound of the covering number, we can obtain an upper bound of $\Lnorm{\sum_\tau \phi_h^\tau \epsilon_h^\tau(\SSS_{h+1}\wV_{h+1})}$. Specifically, by the definition of the $\varepsilon$-cover, we can find $U_\varepsilon\in \cN_{h+1}(\varepsilon)$ such that $\norm{U_\varepsilon-\SSS_{h+1}\wV_{h+1}}_\infty\le \varepsilon$, then
\begin{align}\label{thm1:eqn10}
\begin{aligned}
    &\Lnorm{\sum_\tau \phi_h^\tau \epsilon_h^\tau(\SSS_{h+1}\wV_{h+1})}\\
    &\qquad\le \Lnorm{\sum_\tau \phi_h^\tau \epsilon_h^\tau(U_\varepsilon)}+\Lnorm{\sum_\tau \phi_h^\tau \epsilon_h^\tau(\SSS_{h+1}\wV_{h+1}-U_\varepsilon)}\\
     &\qquad \le \underbrace{\sup_{U'_\varepsilon\in \cN_{h+1}(\varepsilon)}\Lnorm{\sum_\tau \phi_h^\tau \epsilon_h^\tau(U'_\varepsilon)}}_{\textrm{(iii)}}+\underbrace{\Lnorm{\sum_\tau \phi_h^\tau \epsilon_h^\tau(\SSS_{h+1}\wV_{h+1}-U_\varepsilon)}}_{\textrm{(iv)}},
\end{aligned}
\end{align}
where we use triangle inequality in the first step and $U_\varepsilon\in \cN_{h+1}(\varepsilon)$ in the second step. We can thus control $\textrm{(iii)}$ via the classic concentration bound on self-normalized processes in~\citep{jin2021pessimism}, which we write out in \ref{lemma:self-normal} for completion, and have
\begin{align}\label{thm1:eqn11}
    \textrm{(iii)}\le R_\beta \sqrt{2\log(\abs{\cN_{h+1}}H/\delta)+d\log(1+K/\lambda)}
\end{align}
with probability $1-\delta/H$. In addition, for $\textrm{(iv)}$ we have
\begin{align}\label{thm1:eqn12}
    \textrm{(iv)}\le\sum_{\tau=1}^K \norm{\Lambda_{h}^{-1/2}\phi_h^\tau \varepsilon}\le K \norm{\Lambda_h^{-1/2}}_2 \norm{\phi_h^\tau} \cdot\varepsilon\le \frac{\varepsilon K}{\sqrt{\lambda}},
\end{align}
where $\norm{\cdot}_2$ is the spectral norm of the matrix and $\norm{\phi_h^\tau}$ comes from our assumption of linear MDP. Combing \eqref{thm1:eqn9}, \eqref{thm1:eqn10}, \eqref{thm1:eqn11}, \eqref{thm1:eqn12} and $\varepsilon = \sqrt{\lambda}d R_\beta/K$, we have
\begin{align}\label{thm1:eqn13}
\begin{aligned}
     &\Lnorm{\sum_\tau \phi_h^\tau \epsilon_h^\tau(\SSS_{h+1}\wV_{h+1})}\\
     &\qquad \le R_\beta \sqrt{2\log(\abs{\cN_{h+1}}H/\delta)+d\log(1+K/\lambda)} + \frac{\varepsilon K}{\sqrt{\lambda}} \\
     &\qquad \le \sqrt{6}d R_\beta  \sqrt{\log(32c^2e^{4\beta}d^{1/2}K^2\zeta) + \log(H/\delta)+\log(2e^{2\beta}K)}+ d R_\beta\\
     &\qquad \le 4 d R_\beta \sqrt{7\zeta+2\log(c)} \le \frac{c}{2}  dR_\beta\sqrt{\zeta}
\end{aligned}
\end{align}
 with probability $1-\delta/H$. The second step comes from  \eqref{thm1:eqn9}. The third step uses $\log(\zeta)\le \zeta$ and the definition of $\zeta$. The last step holds as long as $c\ge 8\sqrt{7+2\log(c)/\log(2)}$. Consequently, plugging \eqref{thm1:eqn13} into \eqref{thm1:eqn6} shows that
\begin{align*}
    \abs{\BB_h (\SSS_{h+1}\wV_{h+1}) -\wh\BB_h (\SSS_{h+1}\wV_{h+1})}&\le (2+\frac{c}{2})dR_\beta \sqrt{\zeta}\Lnorm{\phi(s,a)}\\
    &\le c d R_\beta\sqrt{\zeta}\Lnorm{\phi(s,a)} = \Gamma_h(s,a)
\end{align*}
holds with probability $1-\delta/H$ if $c\ge 4$. Therefore, $\PP(\cap_{h=1}^H \cE_h)\ge 1-\delta$ and \eqref{thm1:eqn1} ensures that, on the event $\{\cap_{h=1}^H \cE_h\}$, we have
\begin{align*}
     V_1^*(s_1) - \wV_1(s_1)\le \frac{2}{\abs{\beta}} \EE_{\pi^*}[\Gamma_h(s_h,a_h)|s_1] = \frac{e^{\abs{\beta}H}-1}{\abs{\beta}}\cdot 2cd\sqrt{\zeta}\EE_{\pi^*}[\Lnorm{\phi(s_h,a_h)}|s_1].
\end{align*}
Thus completing the proof for Theorem \ref{thm1} for when $\beta > 0$. 

When $\beta<0$, the only difference in the proof is that the function class $\cU_h$ will be replaced by a slightly different function class $\cU_h'$, whose definition we defer to \eqref{l6:eqn0}. As shown in Lemma \ref{lemma:covering-number}, $\cU_h'$ and $\cU_h$ share the same upper bound for the covering number, and thus the above proof remains valid when $\beta<0$ by replacing $\beta$ with $\abs{\beta}$. This completes the proof.
\end{proof}
\section{Proof of Theorem \ref{thm2}}
\label{sec:proofs-for-thm2}
\begin{proof}
Note that the requisite lemmas in Appendix~\ref{sec:proofs-for-thm1} remain valid for Theorem~\ref{thm2}. Thus, to avoid redundancy, we focus on the reference-advantage technique under Assumption~\ref{assump:coverage}. In particular, here we show how the reference-advantage decomposition guides us to a tighter upper bound. 

Without Assumption~\ref{assump:coverage}, the ``good event'' in Lemma~\ref{lemma3} requires a larger uncertainty bonus, $\Gamma_h$. As such, we instead consider the following two events, under which a factor of $\sqrt{d}$ can be shaved off of $\Gamma_h$,
\begin{align*}
    \cE_h&=\left\{\abs{\BB_h  (\SSS_{h+1}\wV_{h+1})(s,a) - \wh\BB_h  (\SSS_{h+1}\wV_{h+1})(s,a) }\le \Gamma_h(s,a)\right\},\\
    \tilde\cE_h&=\left\{\norm{\SSS_{h}\wV_{h} - \SSS_{h}\wV_{h} }_\infty\le R_h\right\}.
\end{align*}
The bonus function now is now $\Gamma_h(s,a) = \gamma\Lnorm{\phi(s,a)}$ with $\gamma = 5 \sqrt{d\zeta}R_\beta$ and $R_h =  \frac{20\sqrt{d\zeta}R_\beta (H+1-h)}{\sqrt{K\kappa}}$ under Assumption~\ref{assump:coverage}. As suggested by \eqref{thm1:eqn1} and \eqref{thm1:eqn2} in the proof of Theorem \ref{thm1}, it is sufficient to show that $1-\PP(\cap_{h=1}^H \cE_h)=\PP(\cup_{h=1}^H \cE_h^c)\le \delta$ with the new bonus function. In particular, we show a stronger result, proving that 
\begin{align}\label{thm2:eqn0}
    \PP\Big((\cup_{h'=h}^{H} \cE_{h'}^c)\cup (\cup_{h'=h}^{H} \tilde\cE_{h'}^c)\Big)\le \delta (H+1-h)/H
\end{align}
for any $h\in[H]$ and the proof is established based on induction from $h=H$ to $h=1$.

\textbf{Base Case $h=H$.}
Our induction starts with the base case $h=H$. Recall that we have shown the upper bounded of the term $\abs{\BB_h (\SSS_{H+1}\wV_{H+1})(s,a) -\wh\BB_h (\SSS_{H+1}\wV_{H+1})(s,a)}$ in \eqref{thm1:eqn6}, by $\wV_{H+1} = V^*_{H+1} =0$ and $\lambda = e^{-2\abs{\beta}}$, \eqref{thm1:eqn6} becomes
\begin{align*}
    &\abs{\BB_H (\SSS_{H+1}\wV_{H+1})(s,a) -\wh\BB_H(\SSS_{H+1}\wV_{H+1})(s,a)}\le 2R_\beta \sqrt{d}\norm{\phi(s,a)}_{\Lambda_H^{-1}}\le\Gamma_H(s,a).
\end{align*}
Therefore, $\PP(\cE_H)=1$. To show that $\tilde \cE_H$ holds with high probability, we again use Lemma \ref{lemma2}, which gives
\begin{align*}
    &\SSS_h V_H^*(s_H) - \SSS_H \wh V_H(s_H) \\
    &\qquad\le \EE_{\pi^*}[\iota_H(s_H,a_H)|s_H] + \EE_{\pi^*}[\SSS_{H+1} V_{H+1}^*(s_{H+1}) - \SSS_{H+1} \wh V_{H+1}(s_{H+1})|s_H]\\
    &\qquad = \EE_{\pi^*}[\iota_H(s_H,a_H)|s_H].
\end{align*}
Conditioned on the event $\cE_H$, we have
\begin{align*}
    \SSS_H V_H^*(s_H) - \SSS_H \wh V_H(s_H) &\le 2\EE_{\pi^*}[\Gamma_H(s_H,a_H)|s_H]=10\sqrt{d\zeta}R_\beta\EE_{\pi^*}[\norm{\phi(s,a)}_{\Lambda_H^{-1}}|s_H]\\
    &\le  \frac{20\sqrt{d\zeta}R_\beta}{\sqrt{K\kappa}}= R_H.
\end{align*}
with probability $1-\delta/H$ given that $K\ge\max\{512\log(2dH/\delta)/\kappa^2,4\lambda/\kappa\}$. The first step comes from Lemma \ref{lemma3}, and the third step uses Assumption \ref{assump:coverage} and Lemma H.5 of \citep{min2021variance}, which we restate Lemma \ref{lemma:full-cover}. Consequently, we have $\PP(\tilde\cE_H^c\cup\cE_H^c)=\PP(\tilde\cE_H^c)\le \delta/H $.

\textbf{Induction from $h+1$ to $h$.}
In this part, we assume that \eqref{thm2:eqn0} holds for $h+1$ and aim to show that it also holds for $h$. We will first show that $\cE_h$ happens with high probability on the event $\tilde\cE_{h+1}$. Inspired by \eqref{thm1:eqn6}, it suffices to consider only $\Lnorm{\sum_\tau \phi_h^\tau \epsilon_h^\tau(\SSS_{h+1}\wV_{h+1})}$. We use the following reference-advantage decomposition to bound this term,
\begin{equation}
    \label{thm2:eqn1}
    \begin{aligned}
        \Lnorm{\sum_\tau \phi_h^\tau \epsilon_h^\tau(\SSS_{h+1}\wV_{h+1})}\le  &\underbrace{\Lnorm{\sum_\tau \phi_h^\tau \epsilon_h^\tau(\SSS_{h+1}V^*_{h+1})}}_{\textrm{(v)}}\\
        &+ \underbrace{\Lnorm{\sum_\tau \phi_h^\tau \epsilon_h^\tau(\SSS_{h+1}\wV_{h+1}-\SSS_{h+1}V^*_{h+1})}}_{\textrm{(vi)}}.
    \end{aligned}    
\end{equation}
As $V_h^*$ is independent of the dataset $\cD$, the term $\textrm{(v)}$ does not depend on the dataset and thus can be bounded Lemma \ref{lemma:self-normal} directly. Specifically, we have, with probability $1-\frac{\delta}{3H}$,
\begin{align}\label{thm2:eqn2}
   \textrm{(v)}\le R_\beta\sqrt{2\log(3H/\delta)+d\log(1+K/\lambda)}\le 2\sqrt{d\zeta}R_\beta.
\end{align}

For the term \textrm{(vi)}, as $\SSS_{h+1}\wV_{h+1}-\SSS_{h+1}V^*_{h+1}$ is correlated with $\cD$, we require uniform concentration analysis similar to that in the proof of Theorem \ref{thm1}. We slightly abuse the notation and use $f_{h+1}$ to denote  $\SSS_{h+1}\wV_{h+1}-\SSS_{h+1}V^*_{h+1}$. For $\beta>0$ and $\beta<0$, we have different function classes for $f_{h+1}$. Similar to Theorem \ref{thm1}, it is sufficient to consider the positive $\beta$ as the different function classes share the same upper bound of the covering number and will not affect the following proof. When $\beta>0$, we can see that $f_{h+1}\in \tilde\cU^*_{h+1}:=\{U_{h+1}-\SSS_{h+1}V^*_{h+1}: U_{h+1}\in\tilde\cU_{h+1}\}$ for $\tilde\cU_{h+1}$ defined in \eqref{thm1:eqn8.5}.

As $\SSS_{h+1}V^*_{h+1}$ is a fixed function of $s$, the $\varepsilon$-covering number of $\tilde\cU^*_{h+1}$ and $\tilde\cU_{h+1}$ are the same, which is $\abs{\cN_{h+1}(\varepsilon)}$ in \eqref{thm1:eqn9}. For notation simplicity, we thus still use $\cN_{h+1}(\varepsilon)$ to denote the $\varepsilon$-cover of $\tilde\cU^*_{h+1}$. By definition, we can find $f_\varepsilon\in \cN_{h+1}(\varepsilon)$ such that $\norm{f_{h+1}-f_\varepsilon}_\infty \le \varepsilon$, which implies that $\norm{f_\varepsilon}_\infty\le R_{h+1}+\varepsilon$ on the event $\tilde\cE_{h+1}$. Consequently, we have
\begin{align*}
    \begin{aligned}
     &\Lnorm{\sum_\tau \phi_h^\tau \epsilon_h^\tau(f_{h+1})}  1\{\norm{f_{h+1}}_\infty\le R_{h+1}\}\\
     &\qquad \le \Lnorm{\sum_\tau \phi_h^\tau \epsilon_h^\tau(f_\varepsilon)}  1\{\norm{f_\varepsilon}_\infty\le R_{h+1}+\varepsilon\} + \Lnorm{\sum_{\tau}\phi_h^\tau \epsilon_h^\tau(f_{h+1}-f_\varepsilon)}.
\end{aligned}
\end{align*}
Similar to \eqref{thm1:eqn11} and \eqref{thm1:eqn12}, with $\varepsilon=\sqrt{\lambda d}R_{h+1}/K$, we have
\begin{align}\label{thm2:eqn4}
\begin{aligned}
        &\Lnorm{\sum_\tau \phi_h^\tau \epsilon_h^\tau(f_{h+1})} 1\{\norm{f_{h+1}}_\infty\le R_{h+1}\} \\
         &\qquad \le(R_{h+1}+\varepsilon) \sqrt{2\log(3H\abs{\cN_{h+1}(\varepsilon)}/\delta)+d\log(1+K/\lambda))}+\frac{\varepsilon K}{\sqrt{\lambda}}\\
         &\qquad \le 2R_{h+1}\sqrt{2\log(\abs{\cN_{h+1}(\varepsilon)})+2\log(3H/\delta)+d\log(2e^{2\beta}K)}+\sqrt{d}R_{h+1}
\end{aligned}
\end{align}
with probability $1-\frac{\delta}{3H}$. As $R_{h+1}/R_\beta =  \frac{20\sqrt{d\zeta}(H-h)}{\sqrt{K\kappa}}\le\frac{20\sqrt{d\zeta}H}{\sqrt{K\kappa}}$ and $\varepsilon=\sqrt{\lambda d}R_{h+1}/K$, given that $K\ge\tilde\Omega(d^2H^2/\kappa)$, the upper bound in \eqref{thm1:eqn9} becomes
\begin{align}\label{thm2:eqn5}
    \begin{aligned}
    \log \abs{\cN_h(\varepsilon)} &\le d\log(1+8 e^{\beta} R_\beta\sqrt{Kd}/\varepsilon )+d\log(1+16 e^{3\beta} R_\beta Kd/\varepsilon )\\
    &\quad +d^2\log(1+8c^2e^{2\beta}R_\beta^2 d^{5/2} \zeta/\varepsilon^2)\\
    &\le d\log(\frac{320e^{2\beta} K^{3/2} H\sqrt{d\zeta}}{\sqrt{K\kappa}})+d\log(\frac{640e^{4\beta} R_\beta K^2 Hd\sqrt{\zeta}}{\sqrt{K\kappa}})\\
    &\quad +d^2\log(\frac{6400c^2e^{4\beta} K^{3/2} Hd^{5/2}\zeta^2}{\sqrt{K\kappa}})\\
    &\le 3d^2\log(\frac{6400c^2e^{4\beta} K^2 Hd^{5/2}\zeta^2}{\sqrt{K\kappa}})\\
    &\le O\left(3d^2\log(6400c^2e^{4\beta} K^2 d^{3/2}\zeta^2)\right)\\
     &\le O\left(3d^2(10\zeta+2\log(c))\right).
    \end{aligned}
\end{align}
Combining \eqref{thm2:eqn4} and \eqref{thm2:eqn5}, we have
\begin{align*}
    \Lnorm{\sum_\tau \phi_h^\tau \epsilon_h^\tau(f_{h+1})} 1\{\norm{f_{h+1}}_\infty\le R_{h+1}\} = O\left(d\sqrt{\zeta}R_{h+1}\right) \le O\left(\frac{20d^{3/2}\zeta H}{\sqrt{K\kappa}}R_\beta\right).
\end{align*}
When $K\ge\tilde\Omega(d^2H^2/\kappa)$, the term is smaller than $\sqrt{d}R_\beta$ and thus non-dominating. In other words, on the event $\tilde\cE_{h+1}=\{\norm{f_{h+1}}_\infty\le R_{h+1}\}$, $\textrm{(vi)}$ in \eqref{thm2:eqn1} is smaller than $\sqrt{d}R_\beta$. Together with \eqref{thm2:eqn2} and \eqref{thm1:eqn6}, we have  
\begin{align*}
    \begin{aligned}
     \abs{\BB_h (\SSS_{h+1}\wV_{h+1})(s,a) -\wh\BB_h (\SSS_{h+1}\wV_{h+1})(s,a)}&\le (2 \sqrt{d}+2\sqrt{d\zeta}+\sqrt{d})R_\beta\Lnorm{\phi(s,a)}\\&\le 5\sqrt{d\zeta}R_\beta \Lnorm{\phi(s,a)}=\Gamma_h(s,a)
\end{aligned}
\end{align*}
with probability at least $1-\frac{2\delta}{3H}$ on the event $\tilde\cE_{h+1}$, which is equivalent to 
\begin{align}\label{thm2:eqn6}
    \PP(\cE_h^c\cap \tilde \cE_{h+1})\le \frac{2\delta}{3H}.
\end{align}
In addition, on event $\cE_h\cap \tilde \cE_{h+1}$, Lemma \ref{lemma2}, Lemma \ref{lemma3}, and Lemma \ref{lemma:full-cover} (which we recall is a restatement of Lemma H.5 of \citep{min2021variance}) jointly imply that
\begin{align*}
    \SSS_h V^*_h(s_h)-\SSS_h \wh V_h(s_h) &\le 2\EE_{\pi^*}[\Gamma_h(s_h,a_h)|s_h]+\EE_{\pi^*}[\SSS_{h+1} V_{h+1}^*(s_{h+1}) - \SSS_{h+1} \wh V_{h+1}(s_{h+1})|s_h] \\
    &\le 10\sqrt{d\zeta}R_\beta\EE_{\pi^*}[\norm{\phi(s_h,a_h)}_{\Lambda_h^{-1}}|s_h]+R_{h+1} \\
    &\le \frac{20\sqrt{d\zeta}R_\beta}{\sqrt{K\kappa}}+R_{h+1} = R_h
\end{align*}
with probability $1-\frac{\delta}{3H}$ for that $K\ge\max\{512\log(6dH/\delta)/\kappa^2,4\lambda/\kappa\}$. Therefore, we have
\begin{align}\label{thm2:eqn7}
    \PP(\tilde\cE_h^c\cap\cE_h\cap \tilde \cE_{h+1})\le \frac{\delta}{3H}.
\end{align}
Using \eqref{thm2:eqn6} and \eqref{thm2:eqn7}, we have
\begin{align}\label{thm2:eqn8}
    \PP\big((\tilde\cE_h^c\cup\cE_h^c)\cap \tilde \cE_{h+1}\big) =\PP(\tilde\cE_h^c\cap\cE_h\cap \tilde \cE_{h+1}) +  \PP(\cE_h^c\cap \tilde \cE_{h+1}) \le \frac{2\delta}{3H} +\frac{\delta}{3H} = \frac{\delta}{H}.
\end{align}
Consequently, by direct calculation, we have
\begin{align}
    \begin{aligned}\label{thm2:eqn9}
       &\hspace{-2em}\PP\Big((\cup_{h'=h}^{H} \cE_{h'}^c)\cup (\cup_{h'=h}^{H} \tilde\cE_{h'}^c)\Big) \\
       =&\PP\Big((\cup_{h'=h+1}^{H} \cE_{h'}^c)\cup (\cup_{h'=h+1}^{H} \tilde\cE_{h'}^c)\Big) + \PP\Big((\tilde\cE_h^c\cup\cE_h^c)\cap \big((\cup_{h'=h+1}^{H} \cE_{h'}^c)\cup (\cup_{h'=h+1}^{H} \tilde\cE_{h'}^c)\big)\Big)\\
       \le &  \PP\Big((\cup_{h'=h+1}^{H} \cE_{h'}^c)\cup (\cup_{h'=h+1}^{H} \tilde\cE_{h'}^c)\Big) +  \PP\big((\tilde\cE_h^c\cup\cE_h^c)\cap \tilde \cE_{h+1}\big) \\
       \le & \frac{\delta(H-h)}{H} + \frac{\delta}{H} = \frac{\delta(H+1-h)}{H}.
\end{aligned}
\end{align}
In the last step of \eqref{thm2:eqn9}, we use \eqref{thm2:eqn8} and the induction assumption that \eqref{thm2:eqn0} holds for $h+1$. 

By induction, \eqref{thm2:eqn9} shows that \eqref{thm2:eqn0} holds for any $h\in[H]$, which implies that $1-\PP(\cap_{h=1}^H \cE_h)=\PP(\cup_{h=1}^H \cE_h^c)\le \delta$ with the bonus parameter $\gamma = 5\sqrt{d\zeta}R_\beta$. The rest of the proof of Theorem~\ref{thm2} then follows that of Theorem~\ref{thm1}, completing the proof.
\end{proof}
\section{Proof of Theorem \ref{thm3}}
\label{sec:proofs-for-thm3}
\begin{proof}
A key technique used by Algorithm~\ref{alg:alg2} is incorporating variance information in estimation. In this section, we highlight how the technique remains viable and beneficial in the risk-sensitive setting. 

Let us recall the ingredients from Appendix~\ref{sec:proofs-for-thm1} and Appendix~\ref{sec:proofs-for-thm2} that we reuse. In Appendix~\ref{sec:proofs-for-thm1}, we have provided the outline for proving the performance of value-iteration style algorithms in risk-averse offline RL in linear MDPs. In Appendix~\ref{sec:proofs-for-thm2}, we show how Assumption~\ref{assump:coverage} and reference-advantage decomposition sharpens the uncertainty bonus $\Gamma_h$. Finally, in this section, we show how incorporating the variance estimator ensures Algorithm~\ref{alg:alg2}'s performance is never worse than that of Algorithm~\ref{alg:alg1} under Assumption~\ref{assump:coverage}.

We start with the consistency of the variance estimator $\wh\sigma_h^2(s,a)$ and then use the Bernstein-type inequality for the self-normalized process to achieve the tighter upper bound in Theorem \ref{thm3}.

\textbf{Consistency of Variance Estimator.}
In this section, we will show that, with probability $1-\frac{\delta}{2H}$,
\begin{align}\label{thm3:eqn1}
    \abs{\sigma_h^2(s_h,a_h)- \wh\sigma_h^2(s_h,a_h)}=\tilde O\left(\frac{dH^2 R_\beta^2}{\sqrt{K\kappa}}\right).
\end{align}
As the clipping at $\underline\sigma^2$ is non-expansive, it is sufficient to show that 
\begin{align}\label{thm3:eqn2}
    \hspace{-0.2cm}\abs{  \underbrace{\{\phi(s,a)^\top \wh\eta_h^{(2)}\}_{[0,(e^{\abs{\beta}H} -1)^2]}-\left(\{\phi(s,a)^\top \wh\eta_h^{(1)}\}_{[0,e^{\abs{\beta}H} -1]}\right)^2}_{\textrm{(vii)}}-\mathrm{Var_h}(\SSS_{h+1}V_{h+1}^*)(s,a)} = \tilde O\left(\frac{dH^2 R_\beta^2}{\sqrt{K\kappa}}\right)
\end{align}
holds with high probability. To prove \eqref{thm3:eqn2}, we first show that $\{\phi^\top \wh\eta_h^{(2)}\}_{[0,R_\beta^2]}-\left(\{\phi^\top \wh\eta_h^{(1)}\}_{[0,R_\beta]}\right)^2$
is close to $\mathrm{Var_h} (\SSS_{h+1}\wh V^{\mathrm{aux}}_{h+1}) = \PP_h ((\SSS_{h+1}\wh V^{\mathrm{aux}}_{h+1})^2) -(\PP_h(\SSS_{h+1}\wh V^{\mathrm{aux}}_{h+1}))^2 $ recalling that $R_\beta: =e^{\abs{\beta}H} -1$. For the estimation of the first moment, that is the term $\PP_h(\SSS_{h+1}\wh V^{\mathrm{aux}}_{h+1})$, the proof is the same as that of Theorem \ref{thm1} for the term $\textrm{(ii)}$ in \eqref{thm1:eqn3}, which gives us
\begin{align}\label{thm3:eqn3}
    \abs{\phi(s,a)^\top \wh\eta_h^{(1)}-\PP_h(\SSS_{h+1}\wh V^{\mathrm{aux}}_{h+1})(s,a)}= \tilde O(d R_\beta)\Lnorm{\phi(s,a)} = \tilde O\left(\frac{dR_\beta}{\sqrt{K\kappa}}\right)
\end{align}
with probability $1-\frac{\delta}{8H}$ given that $K\ge\max\{512\log(16dH/\delta)/\kappa^2,4\lambda/\kappa\}$, where we use Lemma \ref{lemma:full-cover} in the last step. For the second moment, it is sufficient to bound $ \Lnorm{\sum_\tau \phi_h^\tau \epsilon_h^\tau((\SSS_{h+1}\wV^{\mathrm{aux}}_{h+1})^2)}$, which we control by a uniform concentration similar to that in Appendix~\ref{sec:proofs-for-thm1}.

Consider the same $\varepsilon$-cover $\cN_{h+1}(\varepsilon)$ as in the proof of Theorem \ref{thm1}, similar to \eqref{thm1:eqn10} and \eqref{thm1:eqn13}, with $\varepsilon =\sqrt{\lambda}dR_\beta/K$, we have, with probability $1-\frac{\delta}{8H}$,
\begin{align}\label{thm3:eqn4}
     \Lnorm{\sum_{\tau\in\cD^{\rm aux}} \phi_h^\tau \epsilon_h^\tau((\SSS_{h+1}\wV^{\mathrm{aux}}_{h+1})^2)}=\tilde O(dR_\beta^2)   + \frac{2\varepsilon KR_\beta}{\sqrt{\lambda}} =\tilde O(dR_\beta^2),
\end{align}
where the second term in the first equation shares the same spirit of \eqref{thm1:eqn12}. Specifically, for $U_\varepsilon\in \cN_{h+1}(\varepsilon)$ such that $\norm{U_\varepsilon-\SSS_{h+1}\wV_{h+1}}_\infty\le \varepsilon$, we have
\begin{align*}
    \Lnorm{\sum_{\tau\in\cD^{\rm aux}} \phi_h^\tau \epsilon_h^\tau((\SSS_{h+1}\wV_{h+1})^2-U_\varepsilon^2)}\le 2\sum_{\tau=1}^K \norm{\phi_h^\tau}\norm{\Lambda_h^{-1/2}}_2 R_\beta \varepsilon \le \frac{2\varepsilon KR_\beta}{\sqrt{\lambda}},
\end{align*}
where the first step follows the fact that $\abs{a^2-b^2}\le 2\max\{\abs{a},\abs{b}\}\cdot\abs{a-b}$. With \eqref{thm3:eqn4} and Lemma \ref{lemma4}, a standard analysis of ridge regression leads to
\begin{align}\label{thm3:eqn5}
    \abs{\phi(s,a)^\top \wh\eta_h^{(2)}-\PP_h((\SSS_{h+1}\wh V^{\mathrm{aux}}_{h+1})^2)(s,a)} =\tilde O(d R_\beta^2)\Lnorm{\phi(s,a)} =\tilde O\left(\frac{dR_\beta^2}{\sqrt{K\kappa}}\right).
\end{align}
Combing \eqref{thm3:eqn3} and \eqref{thm3:eqn5}, we have, with probability $1-\frac{\delta}{4H}$
\begin{align}
\begin{aligned}\label{thm3:eqn6}
     &\norm{\{\phi^\top \wh\eta_h^{(2)}\}_{[0,(e^{\abs{\beta}H} -1)^2]}-\left(\{\phi^\top \wh\eta_h^{(1)}\}_{[0,e^{\abs{\beta}H} -1]}\right)^2-\mathrm{Var_h} (\SSS_{h+1}\wh V^{\mathrm{aux}}_{h+1})}_\infty\\
     &\qquad  \le\norm{\left(\{\phi^\top \wh\eta_h^{(1)}\}_{[0,e^{\abs{\beta}H} -1]}\right)^2-(\PP_h(\SSS_{h+1}\wh V^{\mathrm{aux}}_{h+1}))^2}_\infty +\norm{\phi^\top \wh\eta_h^{(2)}-\PP_h((\SSS_{h+1}\wh V^{\mathrm{aux}}_{h+1})^2)}_\infty\\
     &\qquad \le2R_\beta\norm{\phi^\top \wh\eta_h^{(1)}-\PP_h(\SSS_{h+1}\wh V^{\mathrm{aux}}_{h+1})}_\infty +\norm{\phi^\top \wh\eta_h^{(2)}-\PP_h((\SSS_{h+1}\wh V^{\mathrm{aux}}_{h+1})^2)}_\infty=\tilde O\left(\frac{dR_\beta^2}{\sqrt{K\kappa}}\right).
\end{aligned}    
\end{align}
Recalling \eqref{thm2:eqn0} in the proof of Theorem \ref{thm2}, we know  $\norm{\SSS_{h+1}\wV^{\mathrm{aux}}_{h+1}-\SSS_{h+1}V^*_{h+1}}_\infty\le R_{h+1}$ with probability $1-\delta$. Replacing $\delta$ by $\frac{\delta}{4H}$, it still holds that, with probability $1-\frac{\delta}{4H}$,
\begin{align*}
    \norm{\SSS_{h+1}\wV^{\mathrm{aux}}_{h+1}-\SSS_{h+1}V^*_{h+1}}_\infty=\tilde O(R_{h+1}) = \tilde O\left(\frac{\sqrt{d}HR_\beta}{\sqrt{K\kappa}}\right).
\end{align*}
Consequently, we have
\begin{align}
    \begin{aligned}\label{thm3:eqn7}
   &\hspace{-2em}\norm{ \mathrm{Var_h} (\SSS_{h+1}\wh V^{\mathrm{aux}}_{h+1})-\mathrm{Var_h}(\SSS_{h+1}V^*_{h+1})}_\infty\\
   &\qquad \le \norm{\PP_h\big((\SSS_{h+1}\wh V^{\mathrm{aux}}_{h+1})^2-(\SSS_{h+1} V^*_{h+1})^2\big)}_\infty+ \norm{\big(\PP_h(\SSS_{h+1}\wh V^{\mathrm{aux}}_{h+1})\big)^2-\big(\PP_h(\SSS_{h+1} V^*_{h+1})\big)^2}_\infty\\
   &\qquad  \le 2R_\beta\norm{\SSS_{h+1}\wh V^{\mathrm{aux}}_{h+1}-\SSS_{h+1} V^*_{h+1}}_\infty+ 2R_\beta\norm{\PP_h(\SSS_{h+1}\wh V^{\mathrm{aux}}_{h+1})-\PP_h(\SSS_{h+1} V^*_{h+1})}_\infty\\=&\tilde O\left(\frac{\sqrt{d}HR_\beta^2 }{\sqrt{K\kappa}}\right),
\end{aligned}
\end{align}
with probability $1-\frac{\delta}{4H}$. Combing \eqref{thm3:eqn6} and \eqref{thm3:eqn7} completes the proof of  \eqref{thm3:eqn2}.

\textbf{Bernstein-Type Inequality for Self-Normalized Process.}  A well-known result in RL is that using a Bernstein-type concentration analysis, as opposed to a Hoeffding-type analysis, sharpens the bounds, and we show that the same technique benefits offline risk-sensitive RL in linear MDPs.
Similar to the proof of Theorem \ref{thm1} and \ref{thm2}, suggested by Lemma \ref{lemma4}, it is sufficient to show that
\begin{align}\label{thm3:eqn8}
    \underbrace{\Snorm{\sum_{\tau\in \cD} \frac{\phi_h^\tau }{\wh\sigma_h(s_h^\tau,a_h^\tau)}\tilde\epsilon_h^\tau(\SSS_{h+1}\wV_{h+1})}}_{\textrm{(viii)}} = \tilde O(\sqrt{d}).
\end{align}
where ${\tilde\epsilon_h^\tau(g)} = \frac{g(s_{h+1}^\tau) - \PP_h g(s_h^\tau,a_h^\tau)}{\wh\sigma_h(s_h^\tau,a_h^\tau)}$. Similar to \eqref{thm2:eqn1}, we use the reference-advantage decomposition an obtain
\begin{align*}
    \textrm{(viii)}\le \underbrace{\Snorm{\sum_{\tau\in \cD} \frac{\phi_h^\tau }{\wh\sigma_h(s_h^\tau,a_h^\tau)}\tilde\epsilon_h^\tau(\SSS_{h+1}V^*_{h+1})}}_{\textrm{(ix)}}+\underbrace{\Snorm{\sum_{\tau\in \cD} \frac{\phi_h^\tau }{\wh\sigma_h(s_h^\tau,a_h^\tau)}\tilde\epsilon_h^\tau(\SSS_{h+1}\wV_{h+1}-\SSS_{h+1}V^*_{h+1})}}_{\textrm{(x)}}.
\end{align*}
Following the proof of Theorem \ref{thm2}, we focus on the reference term $\textrm{(ix)}$ first. As $V_{h+1}^*$ is independent of the data, we can apply a Bernstein-style bound, which we detail in Lemma \ref{lemma:bern-self-normal}, to $\textrm{(ix)}$ directly. 

To utilize Lemma \ref{lemma:bern-self-normal}, we need to show that the conditional mean of $\tilde\epsilon_h^\tau(\SSS_{h+1}V^*_{h+1})$ is 0 and bound the magnitude and conditional variance of $\tilde\epsilon_h^\tau(\SSS_{h+1}V^*_{h+1})$. Specifically, we define the $\sigma$-algebra 
\begin{align*}
    \cF_{h,\tau-1} = \sigma\big(\{(s_h^j,a_h^j)\}_{j=1,j\in \cD}^{\tau}\cup \{s_{h+1}^j\}_{j=1,j\in \cD}^{\tau-1}\big).
\end{align*}
As $\wh\sigma_h^2$ is calculated from the dataset $\cD^{\rm aux}$ and therefore independent of $\cD$, $\tilde\epsilon_h^\tau(\SSS_{h+1}V^*_{h+1})$ is $\cF_{h,\tau}$-measurable and
\begin{align}\label{thm3:eqn9.5}
    \EE[\tilde\epsilon_h^\tau(\SSS_{h+1}V^*_{h+1})|\cF_{h,\tau-1}] = \frac{\EE[\SSS_{h+1}V^*_{h+1}(s_{h+1}^\tau)|s_h^\tau,a_h^\tau]-\PP_h (\SSS_{h+1}V^*_{h+1})(s_h^\tau,a_h^\tau)}{\wh\sigma_h(s_h^\tau,a_h^\tau)}=0.
\end{align}
To bound the magnitude of $\tilde\epsilon_h^\tau(\SSS_{h+1}V^*_{h+1})$, we use \eqref{thm3:eqn1} and the assumption on $K$ that $K\ge\tilde\Omega(d^2H^2R_\beta^4/(\kappa\underline\sigma^4))$, which guarantees that the bias in \eqref{thm3:eqn1} is smaller than $\underline\sigma^2/2$ for sufficiently large $K$ and thus
\begin{align}\label{thm3:eqn10}
\frac{1}{2}\sigma_h^2(s_h,a_h)\le\sigma_h^2(s_h,a_h)-\frac{1}{2}\underline\sigma^2\le \wh\sigma_h^2(s_h,a_h)\le \sigma_h^2(s_h,a_h) + \frac{1}{2}\underline\sigma^2\le \frac{3}{2}\sigma_h^2(s_h,a_h)\le\frac{3}{2}R_\beta^2.
\end{align}
Therefore, we have
\begin{align}\label{thm3:eqn11}
    \abs{\tilde\epsilon_h^\tau(\SSS_{h+1}V^*_{h+1})} \le \frac{\abs{\SSS_{h+1}V^*_{h+1}(s') - \PP_h (\SSS_{h+1}V^*_{h+1})(s,a)}}{\sqrt{\frac{1}{2}\max\{\underline{\sigma}^2, \mathrm{Var}_h(\SSS_{h+1}V_{h+1}^*)(s,a)\}}}\le \sqrt{2\xi( \underline{\sigma}^{2})},
\end{align}
where we let
\[
    \xi(\underline{\sigma}^{2}) = \sup_{h,s,a,s'\sim\PP_h(\cdot|s,a)}\frac{(\SSS_{h+1}V^*_{h+1}(s') - \PP_h (\SSS_{h+1}V^*_{h+1})(s,a))^2}{\max\{\underline{\sigma}^2, \mathrm{Var}_h(\SSS_{h+1}V_{h+1}^*)(s,a)\}}.
\]
Using \eqref{thm3:eqn10}, the conditional variance of $\tilde\epsilon_h^\tau(\SSS_{h+1}V^*_{h+1})$ can also be bounded by
\begin{align}\label{thm3:eqn12}
    \mathrm{Var}[\tilde\epsilon_h^\tau(\SSS_{h+1}V^*_{h+1})|\cF_{h,\tau-1}] =\frac{\mathrm{Var}[\SSS_{h+1}V^*_{h+1}(s_{h+1}^\tau)|s_h^\tau,a_h^\tau]}{\wh\sigma_h^2(s_h^\tau,a_h^\tau)}\le \frac{\sigma_h^2(s_h^\tau,a_h^\tau)}{\wh\sigma_h^2(s_h^\tau,a_h^\tau)}\le 2.
\end{align}
Plugging \eqref{thm3:eqn9.5}, \eqref{thm3:eqn11} and \eqref{thm3:eqn12} into Lemma \ref{lemma:bern-self-normal}, with the assumption that $\xi( \underline{\sigma}^{2} )=O(d)$, we have 
\begin{align*}
    \textrm{(ix)} =\tilde O(\sqrt{d})
\end{align*}
holds with probability $1-\frac{\delta}{4H}$ when \eqref{thm3:eqn1} holds. Following the proof of Theorem \ref{thm2}, the analysis of $\textrm{(x)}$ is the same as that of $\textrm{(vi)}$ in \eqref{thm2:eqn1}, in which we use Lemma \ref{lemma:self-normal} and \ref{lemma:covering-number}. With the same analysis, we have
\begin{align*}
    \norm{\SSS_{h+1}\wV_{h+1}-\SSS_{h+1}V^*_{h+1}}_\infty = \tilde O\left(\frac{\sqrt{d}HR_\beta}{\sqrt{K\kappa}}\right), \text{ and } \textrm{(x)} = \tilde O\left(\frac{d^{3/2}HR_\beta}{\sqrt{K\kappa}\underline\sigma}\right)
\end{align*}
with probability $1-\frac{\delta}{4H}$ given that $K\ge\max\{512 R_\beta^4\log(8dH/\delta)/(\underline\sigma^4\kappa^2),4\lambda R_\beta/\kappa\}$. Together with the assumption that $K\ge \tilde\Omega(d^2H^2R_\beta^4/(\kappa\underline\sigma^4))$, we have $\textrm{(x)}=\tilde O(\sqrt{d})$ and thus \eqref{thm3:eqn8} holds for all $h\in[H]$ with probability  at least $1-H\cdot(\frac{\delta}{4H}+\frac{\delta}{4H}+\frac{\delta}{2H})=1-\delta$. In other words, with Lemma \ref{lemma2}, Lemma \ref{lemma3} and Lemma \ref{lemma4}, we have
\begin{align}\label{thm3:eqn14}
    V_1^*(s_1) - V_1^{\wh\pi}(s_1) \le 
    \tilde O\left(\frac{\sqrt{d}}{\abs{\beta}}\right)\sum_{h=1}^H \EE_{\pi^*}\left[\Snorm{\phi(s_h,a_h)}\Big|s_1=s \right],
\end{align}
with probability $1-\delta$. Moreover, from \eqref{thm3:eqn10}, we have $\wh\sigma_h^2\le \frac{3}{2}\sigma_h^2\le \frac{3}{2}R_\beta^2$, which implies that
\begin{align}\label{thm3:eqn15}
    \Snorm{\phi(s_h,a_h)}\le\sqrt{\frac{3}{2}}\norm{\phi(s_h,a_h)} _{(\Sigma_h^*)^{-1}}\le \sqrt{\frac{3}{2}}R_\beta \Lnorm{\phi(s_h,a_h)}.
\end{align}
Plugging \eqref{thm3:eqn15} into \eqref{thm3:eqn14}, the proof of Theorem \ref{thm3} is completed, and we can see that Theorem \ref{thm3} produces a tighter bound compared to Theorem \ref{thm2}.
\end{proof}
\section{Proof of Lemmas for Main Theorems}
\label{sec:additional-lemmas}
\subsection{Proof of Lemma~\ref{lemma1}}
\begin{proof}
By mean value theorem, for any $x,y$ satisfying $0\le y\le x \le H$, 
\begin{align*}
    e^{\beta x} - e^{\beta y} = \beta e^{\beta z}(x-y)
\end{align*}
for some $z\in [x,y]\in [0,H]$. When $\beta > 0$, it implies that $e^{\beta x} - e^{\beta y} \ge \beta(x-y)$, together with $V_1^*(s_1) - V_1^{\wh \pi}(s_1)\ge 0$ from the definition of $V_1^*$, we have
\begin{align*}
    V_1^*(s_1) - V_1^{\wh \pi}(s_1) \le \frac{1}{\beta}(e^{\beta V_1^*(s_1)}-e^{\beta V_1^{\wh \pi}(s_1)})=\frac{1}{\abs\beta}(\SSS_1 V_1^*(s_1)-\SSS_1 V_1^{\wh\pi}(s_1)).
\end{align*}
When $\beta <0$, the mean value theorem gives $e^{\beta x} - e^{\beta y} \le \beta e^{\beta H}(x-y)$ and thus
\begin{align*}
    V_1^*(s_1) - V_1^{\wh \pi}(s_1) \le \frac{e^{-\beta H}}{\beta}(e^{\beta V_1^*(s_1)}-e^{\beta V_1^{\wh \pi}(s_1)})=\frac{1}{\abs\beta}(\SSS_1 V_1^*(s_1)-\SSS_1 V_1^{\wh\pi}(s_1)).
\end{align*}
The proof of Lemma \ref{lemma1} is thus concluded.
\end{proof}
\subsection{Proof of Lemma \ref{lemma2}}
\begin{proof} We divide our proof into two cases.

\textbf{Case 1: $\beta>0$.} Given the definition of $\SSS_h$ and $\iota_h$, multiplying both sides of \eqref{l3.1:eqn5} and \eqref{l3.1:eqn6} by $e^{\beta(h-1)}$, we obtain \eqref{l2:eqn0} and \eqref{l2:eqn0.1} directly. Using them iteratively with $V_{H+1}^*=\wV_{H+1}=V^{\wh\pi}_{H+1}=0$, we have
\begin{align*}
    \SSS_1 V_1^{*}(s_1) - \SSS_1 V^{\wh\pi}_1(s_1) &\le \SSS_1 V_1^{*}(s_1) - \SSS_1 \wV_1(s_1) \\
    &\le \EE_{\pi^*}[\iota_1(s_1,a_1)|s_1] +\EE_{\pi^*}[\SSS_2 V_2^{*}(s_2) - \SSS_2 \wV_2(s_2)|s_1]\\
    &\qquad \vdots\\&\le  \sum_{h=1}^H\EE_{\pi^*}[\iota_h(s_h,a_h)|s_1].
\end{align*}
\textbf{Case 2: $\beta<0$.} Similarly, multiplying both sides of \eqref{l3.1:eqn5} and \eqref{l3.1:eqn6} by $-e^{\beta H}$ leads to the desired results.
This completes the proof.
\end{proof}
\subsection{Proof of Lemma \ref{lemma3}}
\begin{proof} We divide our proof into two cases.

\textbf{Case 1: $\beta>0$.}
For positive $\beta$, we have
\begin{align*}
    e^{\beta(\widehat Q_h+h-1)}(s,a) &= \{e^{\beta(\widehat r_h(s,a)-1)}(\phi(s,a)^\top \wh w_h+e^{\beta h}) -\Gamma_h(s,a)\}_{[e^{\beta(h-1)},e^{\beta H}]}\\
    &= \{e^{\beta(\widehat r_h(s,a)-1)}(\wh\PP_h(\SSS_{h+1}\wV_{h+1})+e^{\beta h}) -\Gamma_h(s,a)\}_{[e^{\beta(h-1)},e^{\beta H}]}
\end{align*}
and 
\begin{align*}
    \iota_h(s,a) = e^{\beta(r_h(s,a)-1)}(\PP_h (\SSS_{h+1}\wV_{h+1})(s,a)+e^{\beta h})- e^{\beta(\widehat Q_h+h-1)}(s,a).
\end{align*}
We show that $\iota_h(s,a)\ge 0$ first. If $e^{\beta(\wh Q_h+h-1)}(s,a)\le e^{\beta(h-1)}$, it is straight forward that $\iota_h(s,a)\ge 0$ as $\wh r_h(s,a)\in[0,1]$ and $\SSS_{h+1}\wh V_{h+1}\ge e^{\beta h}$. Otherwise, by the definition of $\BB_h$ and $\wh\BB_h$ in \eqref{eqn:operator-B}, on the event $\cE_h$, we have
\begin{align*}
    e^{\beta(\widehat Q_h+h-1)}(s,a) &\le e^{\beta(\widehat r_h(s,a)-1)}(\wh\PP_h(\SSS_{h+1}\wV_{h+1})(s,a)+e^{\beta h}) -\Gamma_h(\cdot,\cdot)\\&\le  e^{\beta(r_h(s,a)-1)}(\PP_h (\SSS_{h+1}\wV_{h+1})(s,a)+e^{\beta h}).
\end{align*}
Equivalent, we have $\iota_h(s,a)\ge0$. For the upper bound of $\iota_h$, we have
\begin{align*}
    &e^{\beta(\widehat r_h(s,a)-1)}(\wh\PP_h(\SSS_{h+1}\wV_{h+1})(s,a)+e^{\beta h}) -\Gamma_h(s,a)\\
    &\qquad\le  e^{\beta(r_h(s,a)-1)}(\PP_h (\SSS_{h+1}\wV_{h+1})(s,a)+e^{\beta h}) \\
    &\qquad\le e^{\beta H}
\end{align*}
on $\cE_h$, which leads to
\begin{align*}
    e^{\beta(\widehat Q_h+h-1)}(s,a) \ge e^{\beta(\widehat r_h(s,a)-1)}(\wh\PP_h(\SSS_{h+1}\wV_{h+1})(s,a)+e^{\beta h}) -\Gamma_h(s,a)
\end{align*}
and thus $\iota_h(s,a)\le 2\Gamma_h(s,a)$. 

\textbf{Case 2: $\beta<0$.} Similarly, following the above proof, the proof for negative $\beta$ can be established by
\begin{align*}
    e^{\beta(\widehat Q_h-H)}(s,a) &= \{e^{\beta\widehat r_h(s,a)}(e^{-\beta H}-\wh\PP_h(\SSS_{h+1}\wV_{h+1})) +\Gamma_h(s,a)\}_{[e^{-\beta(h-1)},e^{-\beta H}]}
\end{align*}
and
\begin{align*}
    \iota_h(s,a) = e^{\beta r_h(s,a)}(e^{-\beta H}-\PP_h (\SSS_{h+1}\wV_{h+1})(s,a))- e^{\beta(\widehat Q_h-H)}(s,a).
\end{align*}
This completes the proof.
\end{proof}

\subsection{Proof of Lemma~\ref{lemma4}}
\begin{proof}
    This is a standard result for ridge regression. From the definition of $\wh\PP_h$, we have
    \begin{align*}
        &\PP_h g_{h+1}(s,a) - \wh\PP_h g_{h+1}(s,a)\\
        &\qquad =\phi(s,a)^\top w_h -\phi(s,a)^\top\Sigma_h^{-1}(\Sigma_h-\lambda \cdot I)w_h + \phi(s,a)^\top\Sigma_h^{-1}(\Sigma_h-\lambda \cdot I)w_h\\
        &\qquad \quad -\phi(s,a)^\top\Sigma_h^{-1}\left(\sum_{\tau}\frac{\phi_h^\tau \cdot g_{h+1}(s_{h+1}^\tau)}{\wh\sigma_h^2(s_h^\tau,a_h^\tau)}\right)\\
        &\qquad= \lambda\phi(s,a)^\top \Sigma_h^{-1}w_h + \phi(s,a)^\top\Sigma_h^{-1}\left(\sum_\tau\frac{\phi_h^\tau\cdot (\PP_h g_{h+1}(s_h^\tau,a_h^\tau)-g_{h+1}(s_{h+1}^\tau))}{\wh\sigma_h^2(s_h^\tau,a_h^\tau)}\right),
    \end{align*}
    which implies that
    \begin{align}
            \begin{aligned}\label{l4:eqn1}
       &\abs{\PP_h g_{h+1}(s,a) - \wh\PP_h g_{h+1}(s,a)}\\
       &\qquad \le \lambda \Snorm{w_h}\Snorm{\phi(s,a)} +  \Snorm{\sum_\tau \frac{\phi_h^\tau }{\wh\sigma_h(s_h^\tau,a_h^\tau)}\cdot \tilde\epsilon_h^\tau(g_{h+1})} \Snorm{\phi(s,a)}.
    \end{aligned}
    \end{align}
    As $\norm{w_h}=\norm{\int_\cS g_{h+1}(s')\mu_h(s')ds'}\le R\sqrt{d}$ from the assumptions of linear MDP, we have
    \begin{align*}
        \lambda \Snorm{w_h} = \lambda \norm{\Sigma_h^{-1/2}w_h}\le \lambda \norm{\Sigma_h^{-1/2}}_2\norm{w_h}\le R\sqrt{d\lambda}.
    \end{align*}
    Plugging this inequality into \eqref{l4:eqn1}, we obtain the desired result for $\abs{\PP_h g_{h+1}(s,a) - \wh\PP_h g_{h+1}(s,a)}$. The proof of $\abs{r_h(s,a) - \wh r_h(s,a)}$ can be viewed as the case that $\tilde\epsilon_h^\tau(\cdot)=0$ and is omitted here. This completes the proof.
\end{proof}
\subsection{Proof of Lemma \ref{lemma5}}
\begin{proof} Again we divide the proof into the following 
 two cases.

\textbf{Case 1: $\beta>0$.} By the definition in \eqref{eqn:operator-B}, we have
    \begin{align*}
    &\abs{e^{\beta (r_h(s,a)-1)}(\PP_h  (\SSS_{h+1}\wV_{h+1})(s,a)+e^{\beta h}) - e^{\beta (\wh r_h(s,a)-1)}(\wh \PP_h (\SSS_{h+1}\wV_{h+1})(s,a)+e^{\beta h}) }\\
    &\qquad \le e^{\beta H}\abs{e^{\beta (r_h(s,a)-1)}-e^{\beta (\wh r_h(s,a)-1)}}+\abs{\PP_h (\SSS_{h+1}\wV_{h+1}) -\wh\PP_h (\SSS_{h+1}\wV_{h+1})}\\
    &\qquad \le e^{\beta H} \beta \abs{r_h-\wh r_h}+\abs{\PP_h (\SSS_{h+1}\wV_{h+1}) -\wh\PP_h (\SSS_{h+1}\wV_{h+1})}.
\end{align*}
The first inequality follows $\abs{\PP_h (\SSS_{h+1}\wV_{h+1})(s,a)+e^{\beta h}}\le e^{\beta H}$, $r_h(s,a)\in[0,1]$ and the triangular inequality. The second inequality follows the fact that $\abs{e^{\beta x}-e^{\beta y}}\le \beta \abs{x-y}$ for $\beta >0$ and $-1\le x,y\le0$, which can be obtained by the mean value theorem as in the proof of Lemma \ref{lemma1}.

\textbf{Case 2: $\beta<0$.} The proof of negative $\beta$ is similar with the fact that $\abs{\PP_h (e^{-\beta H}-\SSS_{h+1}\wV_{h+1})(s,a)}\le e^{\beta H}$ and $r_h(s,a)\in[0,1]$.

This completes the proof.
\end{proof}

\subsection{Proof of Lemma~\ref{lemma:coef-upper-bound}}
\begin{proof}
    By the definition of $\wh w_h$ in Algorithm \ref{alg:alg2}, we have
    \begin{align*}
        \norm{\wh w_h} =\norm{\Sigma_h^{-1}\left(\sum_{\tau=1}^K \frac{\phi_h^\tau\cdot\SSS_{h+1}\wV_{h+1}(s_{h+1}^\tau)}{\wh\sigma_h^2(s_h^\tau,a_h^\tau)}\right)}\le \sum_{\tau=1}^K \norm{\Sigma_h^{-1}\phi_h^\tau\cdot\SSS_{h+1}\wV_{h+1}(s_{h+1}^\tau)}.
    \end{align*}
    Note that $\abs{\SSS_{h+1}\wV_{h+1}(s_{h+1}^\tau)}\le R_\beta $, we have 
    \begin{align*}
        \norm{\wh w_h}\le R_\beta \sum_{\tau=1}^K \norm{\Sigma_h^{-1/2}}\Lnorm{\phi_h^\tau}\le \frac{R_\beta}{\sqrt{\lambda}}\sum_{\tau=1}^K \Lnorm{\phi_h^\tau}=\frac{R_\beta}{\sqrt{\lambda}}\sum_{\tau=1}^K \sqrt{(\phi_h^\tau)^\top \Sigma_h^{-1}\phi_h^\tau}.
    \end{align*}
    By the Cauchy-Schwarz inequality, we have
    \begin{align}
        \begin{aligned}\label{l7:eqn2}
            \norm{\wh w_h}\le\frac{R_\beta\sqrt{K}}{\sqrt{\lambda}} \sqrt{\sum_{\tau=1}^K(\phi_h^\tau)^\top \Sigma_h^{-1}\phi_h^\tau}=\frac{R_\beta\sqrt{K}}{\sqrt{\lambda}} \sqrt{\mathrm{Tr}\left(\Sigma_h^{-1}\sum_{\tau=1}^K(\phi_h^\tau)^\top\phi_h^\tau\right)}.
        \end{aligned}
    \end{align}
    \textbf{Weights $\wh w_h$ in Algorithm \ref{alg:alg1}.}
   If  $\wh\sigma_h = 1$, we are using the standard ridge regression without variance information, then $\wh w$ becomes that in Algorithm \ref{alg:alg1}. In this case, $\Sigma_h = \sum_{\tau=1}^K(\phi_h^\tau)^\top\phi_h^\tau +\lambda \cdot I$ and  \eqref{l7:eqn2} leads to
   \begin{align*}
       \norm{\wh w_h}\le \frac{R_\beta\sqrt{K}}{\sqrt{\lambda}} \sqrt{\mathrm{Tr}\left(\Sigma_h^{-1}(\Sigma_h-\lambda\cdot I)\right)} \le \frac{R_\beta\sqrt{K}}{\sqrt{\lambda}} \sqrt{\mathrm{Tr}\left(\Sigma_h^{-1}\Sigma_h\right)} = R_\beta \sqrt{Kd/\lambda}.
   \end{align*}
 \textbf{Weights $\wh w_h$ in Algorithm \ref{alg:alg2}.} In Algorithm \ref{alg:alg2}, as $\wh\sigma_h^2(s_h^\tau,a_h^\tau)\le R_\beta^2$, we have
 \begin{align*}
     \Sigma_h = \sum_{\tau=1}^K(\phi_h^\tau)^\top\phi_h^\tau /\wh\sigma_h^2(s_h^\tau,a_h^\tau)+\lambda \cdot I \gtrsim \sum_{\tau=1}^K(\phi_h^\tau)^\top\phi_h^\tau /R_\beta^2+\lambda \cdot I= \frac{1}{R_\beta^2}\Phi_h+\lambda \cdot I,
 \end{align*}
where $\Phi_h=\sum_{\tau=1}^K(\phi_h^\tau)^\top\phi_h^\tau$. Then, by \eqref{l7:eqn2}, we have
   \begin{align*}
       \norm{\wh w_h}\le \frac{R_\beta\sqrt{K}}{\sqrt{\lambda}} \sqrt{\mathrm{Tr}\left(\left(\frac{1}{R_\beta^2}\Phi_h+\lambda \cdot I\right)^{-1}\Phi_h\right)}\le R_\beta^2 \sqrt{Kd/\lambda}.
   \end{align*}
   The proof of $\wh \theta_h$ is similar. With $r_h(s,a)\in[0,1]$, we have the desired result in Lemma \ref{lemma:coef-upper-bound}. This completes the proof.
\end{proof}

\subsection{Proof of Lemma~\ref{lemma:covering-number}}
\label{proof:lemma-covering-number}
 \begin{proof} We first more formally introduce the corresponding function classes and covering numbers for when $\beta< 0$. As $\wh\BB_h$ will be different for $\beta>0$ and $\beta<0$, we consider two corresponding function classes.

\textbf{Case 1: $\beta>0$.} We consider the function class
\begin{align*}
    &\cU_h(L_\theta,L_w,L_\gamma,\lambda) 
    \\&\qquad = \left\{U_h(s;\theta,w,\gamma,\Sigma):\cS\to [0,R_\beta] \text{ with } \norm{\theta}\le L_\theta, \norm{w}\le L_w, \gamma\in[0, L_\gamma], \Sigma \gtrsim \lambda\cdot I\right\},
\end{align*}
where 
\begin{align*}
    \begin{aligned}
        &U_h(s;\theta,w,\gamma,\Sigma)
        \\&\qquad =\max_{a}\left\{e^{\beta[\{\phi(s,a)^\top\theta\}_{[0,1]}-1]}(\phi(s,a)^\top w +e^{\beta h})-\gamma \sqrt{\phi(s,a)^\top \Sigma^{-1}\phi(s,a)}\right\}_{[e^{\beta(h-1)},e^{\beta H}]}.
\end{aligned}
\end{align*}
\textbf{Case 2: $\beta<0$.} We consider the function class
\begin{align*}
    &\cU'_h(L_\theta,L_w,L_\gamma,\lambda) 
    \\&\qquad = \left\{U'_h(s;\theta,w,\gamma,\Sigma):\cS\to [0,R_\beta] \text{ with } \norm{\theta}\le L_\theta, \norm{w}\le L_w, \gamma\in[0, L_\gamma], \Sigma \gtrsim \lambda\cdot I\right\},
\end{align*}
where 
\begin{align}
    \begin{aligned}\label{l6:eqn0}
        &U'_h(s;\theta,w,\gamma,\Sigma)
        \\&\qquad =\max_{a}\left\{e^{\beta\{\phi(s,a)^\top\theta\}_{[0,1]}}(e^{-\beta H}-\phi(s,a)^\top w )+\gamma \sqrt{\phi(s,a)^\top \Sigma^{-1}\phi(s,a)}\right\}_{[e^{\beta(h-1)},e^{\beta H}]}.
\end{aligned}
\end{align}
Even though the representation of $\cU_h$ and $\cU'_h$ are slightly different.

We now provide bounds for the covering numbers, again dividing the proof into two cases.

\textbf{Case 1: $\beta>0$}. For any two functions $U_1$ and $U_2$ from $\cU$, let them take parameters $(\theta_1,w_1,\gamma_1,\Sigma_1)$ and $(\theta_2,w_2,\gamma_2,\Sigma_2)$, respectively. As $\{\cdot\}_{a,b}$ and $\max_a$ are contraction map, we have
\begin{align}
\begin{aligned}\label{lcover:eqn1}
    \norm{U_1-U_2}_\infty 
    \le& \underbrace{\sup_{\norm{\phi}\le 1}\abs{e^{\beta[\{\phi^\top\theta_1\}_{[0,1]}-1]}(\phi^\top w_1 +e^{\beta h})-e^{\beta[\{\phi^\top\theta_2\}_{[0,1]}-1]}(\phi^\top w_2 +e^{\beta h})}}_{(a)}\\&+\sup_{\norm{\phi}\le 1}\abs{ \sqrt{\gamma_1^2\phi^\top \Sigma_1^{-1}\phi}-\sqrt{\gamma_2^2\phi^\top \Sigma_2^{-1}\phi}}\\  
    \le& (L_w+e^{\beta h})\beta \sup_{\norm{\phi}\le 1}\abs{\phi^\top\theta_1-\phi^\top\theta_2} +  \sup_{\norm{\phi}\le 1} \abs{\phi^\top w_1-\phi^\top w_2} \\
    &+  \sup_{\norm{\phi}\le 1}\abs{ \sqrt{\gamma_1^2\phi^\top \Sigma_1^{-1}\phi}-\sqrt{\gamma_2^2\phi^\top \Sigma_2^{-1}\phi}}\\
    \le&\beta (L_w+e^{\beta H})\norm{\theta_1-\theta_2} + \norm{w_1-w_2} + \sqrt{\norm{\gamma_1^2\Sigma_1^{-1}-\gamma_2^2\Sigma_2^{-1}}_{F}}.
\end{aligned}
\end{align}
The second step follows from $\abs{\phi^\top w_1+e^{\beta h}}\le L_w + e^{\beta H}$, $e^{\beta[\{\phi^\top\theta_1\}_{[0,1]}-1]}\le1$ and the fact that $\abs{e^{\beta x}-e^{\beta y}}\le \beta \abs{x-y}$ for $\beta >0$ and $-1\le x,y\le0$. Let $\mathcal{C}_w$ be the $\varepsilon/4$-cover of $\{w\in\RR^d:\norm{w}\le L_w\}$, $\mathcal{C}_\theta$ be the $\varepsilon/(4\beta (L_w+e^{\beta H}))$-cover of $\{\theta\in \RR^d:\norm{\theta}\le L_\theta\}$ and $\mathcal{C}_A$ be the  $\varepsilon^2/4$-cover of $\{A\in \RR^{d\times d}:\norm{A}_F\le d^{1/2}L_\gamma^2\lambda^{-1}\}$. As we have $\norm{\gamma^2\Sigma^{-1}}_F\le d^{1/2}L_\gamma^2\lambda^{-1}$, we can see that $\mathcal{C}_w$, $\mathcal{C}_\theta$ and $\mathcal{C}_A$ provide a $\varepsilon-$cover of $\cU_h$. Together with Lemma \ref{lemma11} for the covering number of the Euclidean ball, we have
\begin{align*}
    \log \cN(\varepsilon) \le d\log(1+8L_w /\varepsilon) + d\log(1+8\beta L_\theta (L_w+e^{\beta H})/\varepsilon) + d^2\log(1+8d^{1/2}L_\gamma^2/(\lambda \varepsilon^2)).
\end{align*}
\textbf{Case 2: $\beta<0$}. In the proof of negative $\beta$, the term $(a)$ in \eqref{lcover:eqn1} becomes
\begin{align*}
    \sup_{\norm{\phi}\le 1}\abs{e^{\beta\{\phi^\top\theta_1\}_{[0,1]}}(e^{-\beta H}-\phi^\top w_1 )-e^{\beta\{\phi^\top\theta_2\}_{[0,1]}}(e^{-\beta H}-\phi^\top w_2 )}.
\end{align*}
As $\abs{e^{-\beta H}-\phi^\top w_1}\le e^{\abs{\beta}H}+L_w$ and $e^{\beta\{\phi^\top\theta_1\}_{[0,1]}}\le1$, the second step in \eqref{lcover:eqn1} remains valid if we replace $\beta$ by $\abs{\beta}$. Therefore, we will reach the same upper bound for $\cN'_h(\varepsilon)$. This completes the proof.
 \end{proof}
 \section{Additional Technical Lemmas}
In this section, we provide technical lemmas which are widely used in theoretical reinforcement learning. The proofs of them are omitted and we refer interested readers to the cited sources.
 \begin{lemma}[Lemma B.2 of \citep{jin2021pessimism}]\label{lemma:self-normal} Let $f:\cS\to [0,R]$ be any fixed function, for any $\delta\in(0,1)$, we have
\begin{align*}
    \PP\left(\Lnorm{\sum_\tau \phi_h^\tau \epsilon_h^\tau(f)}^2\ge R^2(2\log(1/\delta)+d\log(1+K/\lambda))\right)\le \delta.
\end{align*}
\end{lemma}

 \begin{lemma}[Bernstein-type inequality for self-normalized process in \citep{zhou2021nearly}]\label{lemma:bern-self-normal}
 Let $\{\eta_t\}_{t=1}^\infty$ be a real-valued stochastic process and let $\{\cF_t\}_{t=1}^\infty$ be a filtration such that $\eta_t$ is $\cF_t$-measurable. Let $\{x_t\}_{t=1}^\infty$ be an $\RR^d$-valued stochastic process where $x_t$ is $\cF_{t-1}$ measurable and $\norm{x_t}\le L$. Let $\Lambda_t =\lambda I_d+\sum_{s=1}^t x_s x_s^\top$. Assume that
     \begin{align*}
         \abs{\eta_t}\le R,~\EE[\eta_t|\cF_{t-1}]=0,~\EE[\eta_t^2|\cF_{t-1}]\le\sigma^2.
     \end{align*}
     Then, for any $\delta>0$, with probability at least $1-\delta$, for all $t>0$, we have
     \begin{align*}
         \Lnorm{\sum_{s=1}^t x_s\eta_s}\le 8\sigma\sqrt{d\log\left(1+\frac{tL^2}{\lambda d}\right)\cdot\log\left(\frac{4t^2}{\delta}\right)} + 4R\log\left(\frac{4t^2}{\delta}\right) =\tilde O(\sigma\sqrt{d}+R).
     \end{align*}
 \end{lemma}
\begin{lemma}[Lemma H.5 of \citep{min2021variance}]\label{lemma:full-cover}
     Let $\phi:\cS\times\cA \to \RR^d$ satisfying $\norm{\phi(s,a)}\le C$ for all $(s,a)\in\cS\times\cA$. For any $K>0$ and $\lambda>0$, define $\mathbb{\bar G}_K=\sum_{k=1}^K\phi(s_k,a_k)\phi(s_k,a_k)^\top+\lambda I_d$ where $(s_k,a_k)$'s are i.i.d, samples from some distribution $\nu$ over $\cS\times\cA$. Let $\mathbb{G} = \EE_\nu[\phi(s,a)\phi(s,a)^\top]$. Then, for any $\delta\in(0,1)$, if $K$ satisfies that 
    \begin{align*}
        K\ge \max\left\{512C^4\norm{\mathbb G^{-1}}^2\log\left(\frac{2d}{\delta}\right),4\lambda\norm{\mathbb G^{-1}}\right\},
    \end{align*}
    with probability at least $1-\delta$, it holds simultaneously for all $u\in\RR^d$ that
    \begin{align*} \norm{u}_{\mathbb{\bar G_K}^{-1}}
        \le \frac{2}{\sqrt{K}}\norm{u}_{\mathbb{G}^{-1}}.
    \end{align*}
\end{lemma}
\begin{lemma}[Lemma D.5 of \citep{jin2020provably}]\label{lemma11}
     For any $\varepsilon>0$, the $\varepsilon$-covering number of the Euclidean ball in $\RR^d$ with radius $R > 0$ is upper bounded by $(1 + 2R/\varepsilon)^d$.
\end{lemma}
\section{Numerical Simulations}
For completeness, we examine a variant of the ModelWin MDP introduced in \cite{thomas2016data} to verify theoretical findings. This particular instance contains 3 states ($S_1$, $S_2$, $S_3$) and 2 actions ($a_1$, $a_2$). Each episode starts from the state $S_1$, where the agent must choose between two actions. Action $a_1$ transitions the agent to either $S_2$ or $S_3$ with probability 0.5 to each state. In contrast, the second action, $a_2$, causes the agent to stay at $S_1$ with probability 0.6; otherwise, the agent will transit to $S_2$ or $S_3$ with equal probability. In states $S_2$ and $S_3$, the agent still has two possible actions, but both always produce a deterministic transition back to $S_1$. The rewards are state-dependent, with $S_1$, $S_2$, and $S_3$ yielding 0.5, 1, and 0, respectively. Formally, 
\begin{align*}
    &\mathbb P(S_2|S_1,a_1)=\mathbb P(S_3|S_1,a_1) = 0.5, \quad \mathbb P(S_1|S_1,a_1)=0\\
    &\mathbb P(S_2|S_1,a_2)=\mathbb P(S_3|S_1,a_2) = 0.2, \quad \mathbb P(S_1|S_1,a_2)=0.6\\
    &\mathbb P(S_1|S_3,a)=\mathbb P(S_1|S_2,a) = 1 \text{ for $a=a_1$ or $a_2$}.
\end{align*}
and
\begin{align*}
    &r(S_1,a) = 0.5, r(S_2,a) = 1, r(S_1,a) = 0 \text{ for $a=a_1$ or $a_2$}.
\end{align*}
In this MDP, the expectation of the total reward is consistently $\mathbb E(\sum_{h=1}^H r_h) =0.5H$, indicating that all policies are equally optimal when $\beta=0$. Consequently, all policies have a suboptimality of 0. However, this MDP is intriguing if we consider how policy choices can impact the reward variance. Staying at $S_1$ guarantees a $0.5$ reward, but leaving $S_1$ yields a $0.5$ chance of obtaining either $1$ or $0$. One can imagine that $a_1$ is a more risky action compared to $a_2$ as $a_2$ is more likely to keep the agent at $S_1$. The behavior policy we use to generate the offline data is taking $a_1$ and $a_2$ randomly with equal probability. Considering the entropic risk measure we study, the agent is risk-seeking when $\beta>0$ and risk-averse when $\beta<0$. Therefore, the optimal policy will be consistently taking action $a_1$ when $\beta>0$ and consistently taking $a_2$ when $\beta<0$. 
We take our first algorithm as an example and conduct experiments using the above environment based on this algorithm. 

\begin{figure}[htb]
\includegraphics[width=8cm]{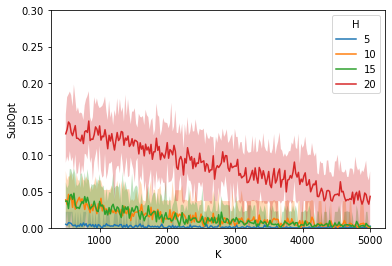}
\includegraphics[width=8cm]{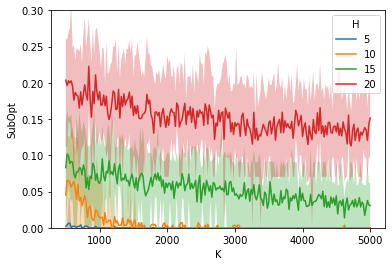}
\centering
\caption{Each panel reports the suboptimality of the learned policy from Algorithm \ref{alg:alg1} for different $K$ and $h$. $\beta =0.5$ (left) and $\beta =1$ (right). The results are averaged over 20
independent trails, and the mean results are plotted as solid lines. The error bar area corresponds to the $80\%$ confidence interval. }\label{fig1}
\end{figure}

We evaluate the scenarios $H=5,10,15,20$ and $\beta = 0.5, 1$ in the experiment. The suboptimality results are reported in Figure \ref{fig1}. We can see that with a larger $K$, the suboptimality goes to $0$, which serves as simulation evidence for our algorithm. Moreover, Figure \ref{fig1} illustrates that with an increase in $H$ and $|\beta|$, there is a corresponding rise in the suboptimality gap, aligning with our theoretical result. 

\end{document}